\documentclass[10pt,twocolumn,letterpaper]{article}

\usepackage[pagenumbers]{cvpr} 
\usepackage[accsupp]{axessibility}

\definecolor{cvprblue}{rgb}{0.21,0.49,0.74}
\usepackage[pagebackref,breaklinks,colorlinks,allcolors=cvprblue]{hyperref}

\def\paperID{8876} 
\def\confName{CVPR}
\def\confYear{2026}

\title{Pseudo-Expert Regularized Offline RL for End-to-End Autonomous Driving \\ in Photorealistic Closed-Loop Environments}

\author{Chihiro Noguchi \qquad Takaki Yamamoto \\
InfoTech,\ Toyota Motor Corporation\\
{\tt\small \{chihiro\_noguchi\_aa, takaki\_yamamoto\}@mail.toyota.co.jp}
}

\begin{document}
\maketitle

\begin{abstract}
End-to-end (E2E) autonomous driving models that take only camera images as input and directly predict a future trajectory are appealing for their computational efficiency and potential for improved generalization via unified optimization; however, persistent failure modes remain due to reliance on imitation learning (IL). While online reinforcement learning (RL) could mitigate IL-induced issues, the computational burden of neural rendering-based simulation and large E2E networks renders iterative reward and hyperparameter tuning costly. We introduce a camera-only E2E offline RL framework that performs no additional exploration and trains solely on a fixed simulator dataset. Offline RL offers strong data efficiency and rapid experimental iteration, yet is susceptible to instability from overestimation on out-of-distribution (OOD) actions. To address this, we construct pseudo ground-truth trajectories from expert driving logs and use them as a behavior regularization signal, suppressing imitation of unsafe or suboptimal behavior while stabilizing value learning. Training and closed-loop evaluation are conducted in a neural rendering environment learned from the public nuScenes dataset. Empirically, the proposed method achieves substantial improvements in collision rate and route completion compared with IL baselines. Our code is available at \url{https://github.com/ToyotaInfoTech/PEBC}.
\end{abstract}

\begin{figure}[t]
  \centering
   \includegraphics[width=0.98\linewidth]{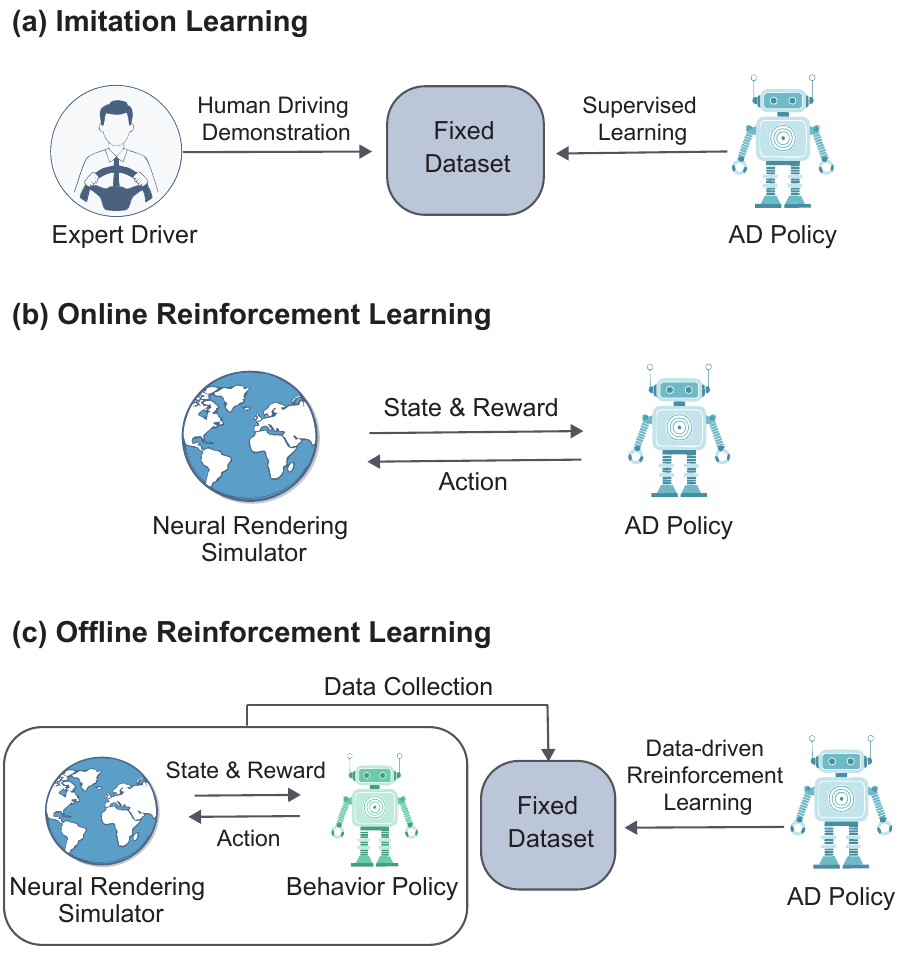}
   \caption{Driving policy learning paradigms. (a) Imitation Learning (IL): Supervised learning on a fixed expert dataset. (b) Online Reinforcement Learning (RL): Policy learns via continuous live interaction with a simulator. (c) Offline Reinforcement Learning (Our Approach): Policy learns from a fixed, pre-collected dataset without new simulator interaction.}
   \label{fig:intro}
\end{figure}

\section{Introduction}
\label{sec:intro}

Recent advances in deep learning have spurred interest in end-to-end (E2E) autonomous driving methods that take only camera images and directly output future trajectories (or waypoints). Such methods promise unified optimization of perception and control, improved generalization, and higher inference throughput per compute. Most existing approaches rely on imitation learning (IL), regressing or classifying expert actions from large driving logs. However, IL suffers from intrinsic limitations: (1) \textbf{Covariate shift / out-of-distribution (OOD) states}: Errors accumulate when the policy encounters states not represented in expert data, leading to compounding failures. (2) \textbf{Causal confusion}: Models may latch onto spurious correlations (e.g., a lead car's brake lights) rather than true causal signals (e.g., the obstacle causing the braking)~\cite{de2019causal}.

Reinforcement learning (RL) can, in principle, mitigate these issues by optimizing expected long-term returns. Yet applying \textit{online} RL to E2E driving faces major barriers: (a) the safety and sample cost of large-scale exploration in real vehicles or high-fidelity simulators, (b) heavy computation from photorealistic neural rendering environments, and (c) costly iterative reward and hyperparameter tuning for E2E models. These factors slow practical R\&D cycles.

\textit{Offline} RL (Fig.~\ref{fig:intro}c) addresses these constraints by learning policies purely from fixed datasets without additional environment interaction—a paradigm highly relevant for domains where exploration is unsafe or expensive~\cite{levine2020offline,chen2019top,jaques2019way}. In the autonomous driving context, a core difficulty is Q-function overestimation on actions not supported by the data (OOD actions), which can induce unsafe policy updates. Mitigation strategies
fall broadly into (i) constraining the learned policy to remain within (or close to) the behavior policy’s support~\cite{fujimoto2021minimalist,park2025flow}, and (ii) pessimistically biasing value estimation~\cite{kumar2020conservative,yu2020mopo}.

Despite increasing RL adoption in other fields (e.g., large language model alignment~\cite{shao2024deepseekmath,guo2025deepseek}), applications to E2E autonomous driving remain limited. Pioneering efforts such as RAD~\cite{gao2025rad} construct a 3D Gaussian Splatting simulator with large in-house data to demonstrate the efficacy of online RL, while ReCogDrive~\cite{li2025recogdrive} leverages NAVSIM~\cite{dauner2024navsim} with GRPO-based training. However, systematic validation of a purely offline setting has not been reported.

We propose a camera-only E2E offline RL framework that forgoes new online exploration, instead leveraging pre-collected simulator data. Our method employs a behavior-regularized actor-critic formulation. While standard Behavior Cloning (BC) would mimic the behavior policy---including its suboptimal or unsafe actions---our approach introduces a key modification: we construct \textbf{pseudo ground-truth (expert-like) trajectories} from expert driving logs. This signal is used as a constraint to suppress the imitation of unsafe behavior while stabilizing value learning. To ensure reproducibility, all training and closed-loop evaluation occurs in a neural rendering environment built from the public nuScenes dataset.

To our knowledge, this work is among the first to systematically evaluate a camera-only E2E offline RL policy in closed-loop form. Comprehensive ablations validate the importance of our proposed regularization, reward shaping, and behavior policy selection. Our final model jointly demonstrates a substantial reduction in collision rate and improved route completion compared to IL-based methods.

Our contributions are summarized as follows.
\begin{itemize}
    \item We propose a camera-only, E2E offline RL framework for autonomous driving, which learns from a fixed dataset without costly online environment interaction.
    \item We introduce a pseudo-expert regularization technique using interpolated ground-truth (GT) expert trajectories to stabilize training and avoid cloning unsafe behaviors.
    \item Our method substantially improves safety and efficiency over IL baselines, and we provide a detailed empirical analysis of the impact of behavior policy composition on performance.
\end{itemize}

\section{Related Work}
\label{sec:related_works}

\subsection{E2E Autonomous Driving}

E2E autonomous driving involves generating control outputs directly from raw sensor inputs. Recent paradigms improve planning by jointly learning multiple perception tasks within a unified architecture~\cite{hu2023planning,jiang2023vad}. More recently, vision-language models (VLMs) pre-trained on web-scale datasets have been leveraged to enhance generalization and robustness in novel environments~\cite{jiang2024senna,renz2025simlingo}. Many of these approaches use a continuous action space, modeling expert trajectories with direct regression~\cite{sun2025sparsedrive,weng2024drive,jia2025drivetransformer} or generative policies like diffusion models~\cite{liao2025diffusiondrive,zheng2024genad}. In contrast, discrete-action formulations like VADv2~\cite{chen2024vadv2} and MDP~\cite{li2024hydra} create a finite action vocabulary using k-means clustering, thereby framing policy learning as a classification problem. In this study, we adopt a discrete action space and proposes a model based on VADv2.

\subsection{RL for Autonomous Driving}

Most RL research for autonomous driving has been conducted in physics-based simulators such as CARLA~\cite{dosovitskiy2017carla,toromanoff2020end,liang2018cirl}. Major directions include: hybrid schemes that initialize with IL and then improve with RL or alternate IL/RL phases~\cite{liang2018cirl,lu2023imitation}; model-based approaches that leverage a learned world model to boost sample efficiency~\cite{li2024think2drive,hu2024solving,chen2021learning,chen2021interpretable}; knowledge distillation setups where an RL-trained teacher with privileged information guides a student via imitation~\cite{zhang2021end,li2024think2drive}; and automation of reward / evaluation metric design using large vision–language models (VLMs)~\cite{ye2025lord,huang2025vlm}; or learning state-based policies via massive-scale self-play~\cite{cusumano2025robust}. Recently, end-to-end policy optimization has gained traction, exemplified by ReCogDrive~\cite{li2025recogdrive} (NAVSIM~\cite{dauner2024navsim} + GRPO~\cite{shao2024deepseekmath}) and RAD~\cite{gao2025rad} (3DGS~\cite{kerbl20233d} environment + PPO~\cite{schulman2017proximal}). Our proposed method is based on offline RL---to the best of our knowledge, this is the first work to apply it to E2E autonomous driving.

\subsection{Offline RL}

A core difficulty in offline RL is the propensity of value estimators to overestimate returns for OOD actions. To mitigate this extrapolation error, existing methods constrain policy updates to remain within the empirical state–action support. Broad methodological families include: (i) behavioral regularization~\cite{nair2020awac,fujimoto2021minimalist,tarasov2023revisiting,wang2022diffusion,park2025flow}; (ii) Q-function conservation~\cite{kumar2020conservative}; (iii) in-sample learning~\cite{kostrikov2021offline,xu2023offline,garg2023extreme,hansen2023idql}; (iv) model-based offline RL with uncertainty-aware pessimism~\cite{yu2020mopo,kidambi2020morel,an2021uncertainty,yu2021combo}; and (v) generative modeling to guide policy improvement~\cite{chen2021decision,janner2021offline,janner2022planning}. Our approach employs behavioral regularization using pseudo GT trajectories derived from expert driving data.

\section{Preliminaries}
\label{sec:prelimnaries}

\textbf{Reinforcement Learning.}
RL formally addresses goal-directed learning from interaction. The standard framework for an RL problem is the Markov decision process, which is specified by a tuple $(\mathcal{S}, \mathcal{A}, \mathcal{R}, \mathcal{P}, \gamma)$. Here, $\mathcal{S}$ is the state space, $\mathcal{A}$ is the action space, $\mathcal{R}: \mathcal{S}\times \mathcal{A}\to \mathbb{R}$ is the reward function, $\mathcal{P}:\mathcal{S}\times \mathcal{A}\to \Delta(\mathcal{S})$ is the state transition dynamics, and $\gamma\in [0,1)$ is the discount factor. $\Delta(\mathcal{X})$ denotes the set of probability distributions on space $\mathcal{X}$. 

A policy $\pi:\mathcal{S}\to\Delta(\mathcal{A})$ defines a probability distribution over actions given a state. The goal of RL is to learn a policy that maximizes the expected cumulative reward: $J(\pi)=\mathbb{E}_{\pi} \left[\sum_{t=0}^\infty \gamma^{t} r(s_t, a_t) \right]$, where $r \in \mathcal{R}$. The Q-function is defined as the expected cumulative reward when taking action $a$ in state $s$ and following policy $\pi$ thereafter:

$$Q^\pi(s, a) = \mathbb{E}_{\pi} \left[ \sum_{t=0}^{\infty} \gamma^t r(s_t, a_t) \mid s_0 = s, a_0 = a \right].$$

For end-to-end autonomous driving, the state $s_t\in\mathcal{S}$ is a composite of two data modalities. It includes exteroceptive sensor data, such as multi-camera images for perceiving the environment, and proprioceptive data, such as vehicle speed and steering angle from the CAN bus, which describes the ego-vehicle's internal state.

\noindent
\textbf{Offline Reinforcement Learning.}
In offline RL, a policy is learned using only a pre-collected fixed dataset $\mathcal{D}=\{(s_i,a_i,r_i,s_i^\prime)\}_{i=1}^{N}$ without any additional data collection. This dataset is assumed to be collected by a behavior policy $\pi_\beta$. Unlike online RL, no new data can be gathered through environment interaction during training.

\begin{figure*}[t]
  \centering
   \includegraphics[width=0.98\linewidth]{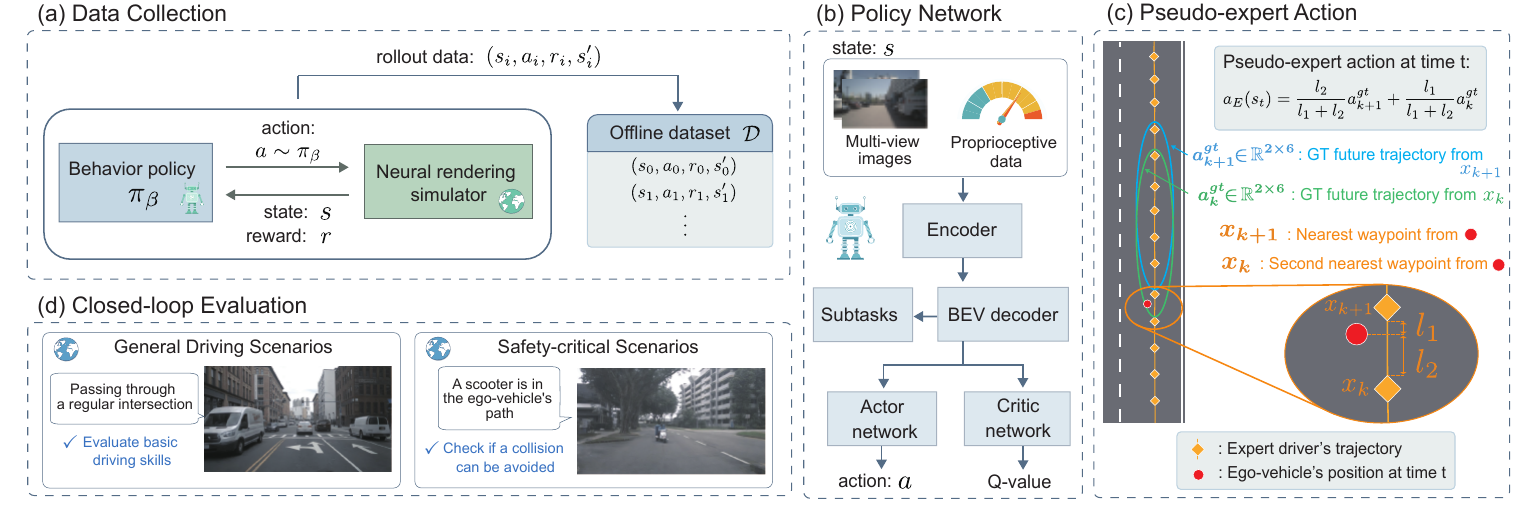}
   \caption{An overview of the proposed camera-only offline RL framework. (a) Data Collection: An offline dataset is generated by executing various behavior policies in a neural rendering simulator to collect rollout data. (b) Policy Network: A discrete action-space Actor-Critic network , built upon an Encoder and BEV Decoder, is trained using the fixed offline dataset. (c) Pseudo-expert Action: A pseudo-expert action is generated from the offline dataset by interpolating expert ground-truth trajectories. (d) Closed-loop Evaluation: The trained policy is evaluated in the simulator on two distinct suites: General Driving Scenarios and Safety-critical Scenarios.}
   \label{fig:overview}
\end{figure*}

\section{Proposed Approach}
\label{sec:proposed_approach}

We propose an offline RL framework for end-to-end autonomous driving models that use only camera images as input (Fig.~\ref{fig:overview}). Our approach adopts a discrete action space Actor-Critic architecture based on VADv2, which is trained with a behavioral cloning (BC) regularizer to ensure stability.
Crucially, we use pseudo GT trajectories generated from expert driving logs to avoid imitating unsafe actions (e.g., collisions) present in the behavior policy.

\subsection{Policy Network}

Our policy network consists of four main components: an encoder, a BEV decoder, an actor network, and a critic network.

\noindent
\textbf{Encoder.} The encoder $f_\text{enc}$ takes multi-view camera images and ego-vehicle proprioceptive information as input, and extracts a unified state representation $z_\text{img} = f_\text{enc}(s_t)$. 

\noindent
\textbf{BEV Decoder.} The BEV decoder transforms the perspective-view image features into a spatially coherent bird's-eye-view representation $h_{\rm bev}\in \mathbb{R}^{H\times W \times D}$. Learnable BEV queries $\{q_\text{bev}^i\}_{i=1}^{H\times W} \in \mathbb{R}^{H\times W \times D}$ attend to the image features $z_\text{img}$ via deformable cross-attention (DCA)~\cite{li2024bevformer,zhu2021deformable} to make the BEV feature map  $z_\text{bev} \in \mathbb{R}^{H\times W \times D}$, with $H$ and $W$ denoting the BEV height and width and $D$ the hidden dimension.

\noindent
\textbf{Actor Network.} The actor network $\pi_\theta(a|s_t)$ outputs a categorical probability distribution over a discrete action set $\mathcal{A} = \{a_1, a_2, \ldots, a_K\}$ of size $K$. Each discrete action $a_k$ represents a future ego-vehicle trajectory obtained by clustering expert trajectories via k-means. Following prior methods, the model first uses learnable agent and map queries to attend to the BEV features $z_\text{bev}$, promoting a structured understanding of dynamic objects and the static environment.
The policy is then formed using $K$ learnable action queries $\{q_k^\text{act}\}_{k=1}^{K} \in \mathbb{R}^{K \times D}$ that attend to the BEV, agent, and map features through DCA. The attended feature for  action $k$ is represented by $h_k \in \mathbb{R}^{D}$. The action logits are then computed via a MLP head:

$$\pi_\theta(a_k|s_t) = \frac{\exp(\text{MLP}_\text{act}(h_k))}{\sum_{j=1}^{K} \exp(\text{MLP}_\text{act}(h_j))}.$$

\noindent
\textbf{Critic Network.} The critic network estimates Q-values $Q_\psi(s_t, a_k)$ for each state–action pair. In our implementation, we adopt a shared-encoder design where the critic directly outputs $K$ Q-values from the corresponding action features $h_k$, enabling efficient computation during both training and inference.

\subsection{Actor-Critic Training}

\noindent
\textbf{Critic Loss.}
The critic network is trained to approximate the expected return by minimizing the one-step temporal difference (TD) error with target network stabilization. The loss function is defined as:

$$\mathcal{L}_\text{critic}(\psi) = \mathbb{E}_{(s,a,r,s^\prime) \sim \mathcal{D}, a^\prime\sim\pi_{\theta^\prime}(\cdot|s^\prime)} \left[ \left( Q_\psi(s, a) - y \right)^2 \right],$$

\noindent
where the target value $y = r + \gamma (1-d) Q_{\psi^\prime}(s^\prime, a^\prime).$ Here, $d$ is an indicator for terminal states ($d=1$ if $s'$ is terminal, $0$ otherwise). The target networks' parameters, $\theta'$ and $\psi'$, are periodically updated using an exponential moving average (EMA) of the online network parameters.

\noindent
\textbf{Actor Update with Pseudo-expert Regularization.}
The actor network $\pi_\theta(a|s)$ is updated to maximize the expected Q-values of its selected actions. A critical challenge in offline RL is that maximizing $ Q_\psi(s, a)$ can lead the policy to exploit OOD actions where the critic produces erroneously high Q-values~\cite{fujimoto2019off}. To counteract this, we introduce a BC loss that regularizes the policy.

To address this, we introduce a pseudo-expert regularization scheme. Instead of cloning the behavior policy, we guide the actor towards a pseudo-expert policy $\pi_E$ derived from a separate, clean dataset of GT expert trajectories. For each state $s$ encountered during training, we dynamically construct its corresponding pseudo-expert action $a_E(s)$.

As illustrated in Fig.~\ref{fig:overview}c, this process begins by identifying the two GT waypoints nearest to the ego-vehicle's position in state $s$. The expert trajectories associated with these waypoints are then linearly interpolated to generate a synthetic reference trajectory tailored to the vehicle's precise state. Finally, the discrete action prototype in our vocabulary $\mathcal{A}$ with the minimum Euclidean distance to this reference trajectory is designated as the pseudo-expert action $a_E(s)$. The actor is then updated with the off-policy policy gradient method~\cite{levine2020offline,degris2012off}. The final actor loss is defined as:

\begin{align}
\mathcal{L}_\text{actor}(\theta) &= \underbrace{\mathbb{E}_{s \sim \mathcal{D}, a^\pi\sim \pi_\theta(\cdot|s)} \left[A^{\pi_\theta}(s,a^\pi)\log\pi_\theta(a^\pi|s) \right]}_{\text{RL Objective}} \notag \\
& \quad \underbrace{-\alpha\mathbb{E}_{s\sim\mathcal{D}}\left[ \log\pi_\theta(a_E(s)|s) \right]}_{\text{Pseudo-Expert BC}},
\label{eq:actorloss}
\end{align}

\noindent
where $A^{\pi}(s,a)=Q^{\pi_\theta}(s,a)-\sum_{a^\prime\in\mathcal{A}}\pi_\theta(a^\prime|s)Q^{\pi_\theta}(s,a^\prime)$ is the advantage function, and $\alpha$ is a hyperparameter that balances the RL objective with the pseudo-expert regularization. 
For the RL objective term in Eq.~\ref{eq:actorloss}, we replace the expectation over the current policy's state distribution commonly deployed in on-policy setups with an empirical average over states sampled from the offline dataset $\mathcal{D}$. This off-policy estimation approach is widely adopted in offline RL~\cite{levine2020offline,degris2012off} to approximate policy gradients without environment interaction. The selection of an appropriate behavior policy is generally non-trivial. We therefore conduct an empirical analysis to examine the impact of different behavior policy choices on model performance.

\subsection{Reward Design}
The design of the reward function is critical for guiding the RL agent toward desirable driving behaviors. A well-shaped reward function should encourage safe and efficient navigation while penalizing actions that lead to failure. Our total reward $r$ is a weighted sum of several components designed to capture different aspects of the driving task:
$r = w_\text{imitation} r_\text{imitation} + w_\text{event} r_\text{event},$
where $w_\text{imitation}$ and $w_\text{event}$ are weighting coefficients that balance the contribution of each term. Each component is detailed below.

\noindent
\textbf{Expert Imitation ($r_\text{imitation}$).} 
To provide a dense learning signal, we incorporate a reward based on proximity to the pseudo-expert trajectory. This component guides the agent toward expert-like behavior at each step. The expert imitation reward $r_\text{imitation}$ is defined as the negative distance between the behavior policy's action and the pseudo-expert action, ensuring that closer alignment results in a higher reward:
$r_\text{imitation} = -D(a^{\pi_\beta}_t,  a_E(s_t)),$
where $a^{\pi_\beta}_t$ is the action from the behavior policy at timestep $t$, $a_E(s_t)$ is the corresponding pseudo-expert action, and $D(\cdot, \cdot)$ is the mean squared Euclidean distance between the two trajectories.

\noindent
\textbf{Terminal Event Penalties ($r_\text{event}$).} 
To strongly discourage catastrophic failures, we apply large, sparse penalties for critical events that terminate an episode. These events include collisions with other agents (e.g., vehicles, pedestrians) or static objects, and driving off the designated road area or route. 
The event penalty $r_\text{event}$ is $0$ by default, but set to a large negative constant ($C_\text{collision}$, $C_\text{off-road}$, or $C_\text{off-route}$) if a collision, off-road, or off-route event occurs, respectively. These penalties provide a strong discouraging signal against unsafe actions. This sparse and dense reward combination balances expert-following with critical error avoidance.

\section{Experiments}
\label{sec:experiments}
We experimentally validate our proposed offline RL framework in a closed-loop simulation environment against IL baselines. Our analysis includes a main comparison, detailed ablation studies, and qualitative examples.

\subsection{Experimental Setup}

\noindent
\textbf{Dataset and Simulator.}
All experiments are conducted in the NeuroNCAP~\cite{ljungbergh2024neuroncap} simulator, a photorealistic, closed-loop evaluation platform that uses the NeuRAD~\cite{tonderski2024neurad} neural rendering engine built on the nuScenes dataset~\cite{caesar2020nuscenes}. We generate our offline dataset by running a mix of behavior policies ($\pi_\beta$) in NeuroNCAP. The resulting dataset contains a diverse distribution of outcomes, including successful navigation, collisions, and off-route events, providing the necessary signal for robust value learning. The pseudo-expert trajectories used for regularization are derived directly from the original GT nuScenes logs. Our implementation extends the original NeuroNCAP environment to include collision detection with road boundaries, pedestrians, and barriers. See Appendix~\ref{sec:appendix_data_collection} for data collection details.

\noindent
\textbf{Behavior Policies.}
To generate datasets for our experiments, we use a collection of policies with varying levels of expertise and stochasticity. This allows us to study the impact of dataset composition on performance (Sec.~\ref{sec:ablation}). The policies used to generate trajectories are:

\noindent
\underline{Noisy Imitation Policies}: We use VAD and VADv2 models, pre-trained via IL, as our base policies. To generate exploratory data, we create stochastic versions of these policies by adding zero-mean Gaussian noise to the predicted waypoints. We use three noise levels $\sigma\in\{0.1, 0.2, 0.4\}$ for VAD and VADv2 models.

\noindent
\underline{Random Policy}: To ensure broad coverage of the action space and provide clear examples of suboptimal behavior for value learning, we include a uniform random policy. This policy selects an action prototype from the vocabulary $\mathcal{A}$ with equal probability at each timestep.

Our final offline datasets, used in Table~\ref{tab:main_results}, are generated either by these policies individually or by specific mixtures (e.g., VAD($\sigma=0.2$) + VAD($\sigma=0.4$)). 
This approach provides a rich mix of exploratory actions and diverse failure cases, which is critical for robust offline RL training.

\subsection{Evaluation Metrics}
\label{subsec:metrics}

To assess policy performance, we conduct closed-loop evaluations in the NeuroNCAP simulator. We measure performance using three primary metrics common in prior work (CR, RC, and Jerk) and introduce a unified metric (SRC and JSR) to capture the trade-off between progress and safety across two distinct scenario suites.

\noindent
\textbf{Evaluation Scenarios.}
    

\noindent
\underline{General Driving (nuScenes Validation)}: This standard benchmark consists of 137 scenes from the nuScenes validation split, excluding those where the ego-vehicle is stationary. It is used to evaluate the policy's general driving competency in normal traffic situations.

\noindent
\underline{Safety-Critical Scenarios}: This suite, provided by NeuroNCAP, includes 20 challenging scenes designed to test the policy's robustness to sudden hazards. These scenarios involve an adversarial vehicle approaching from the side or front, or stopping in the ego-vehicle's path, evaluating the policy's ability to execute safe avoidance maneuvers. For these scenarios, we fixed the navigation command to ``straight'' to prevent the ego-vehicle from being forced to return to its route while attempting to avoid a collision.
    
\noindent
\textbf{Primary Metrics.}

    
    
    


\noindent
\underline{Collision Rate (CR)}: The percentage of evaluation episodes that terminate due to a collision with another agent (e.g., vehicle, pedestrian), or a static map object.

\noindent
\underline{Route Completion (RC)}: The average percentage of the planned route completed per episode.

\noindent
\underline{Jerk}: The time-averaged rate of change of the ego-vehicle's acceleration (m/s$^3$). This metric quantifies ride comfort and smoothness, with lower values being preferable.

\noindent
\underline{Safety-Weighted Route Completion (SRC)}: To evaluate the critical trade-off between efficiency in normal driving and robustness in hazardous situations, we introduce this unified metric: $ \text{SRC} = \text{RC}_{\text{Gen}} \times (1 - \text{CR}_{\text{Safe}}), $
    where $\text{RC}_{\text{Gen}}$ is the Route Completion from the General Driving suite, and $\text{CR}_{\text{Safe}}$ is the Collision Rate from the Safety-Critical suite.

\noindent
\underline{Joint Safety Rate (JSR)}: We also introduce a metric to evaluate the agent's ability to remain collision-free in both general and critical scenarios: $ \text{JSR} = (1 - \text{CR}_{\text{Gen}}) \times (1 - \text{CR}_{\text{Safe}}), $
    where $\text{CR}_{\text{Gen}}$ is the Collision Rates from the General Driving suites.

\begin{table*}[ht]
\centering
\caption{\textbf{Main Results on Closed-Loop Evaluation.} We compare our offline RL method against IL baselines on the General Driving and Safety-Critical scenario suites. The symbols * and $\dag$ denote the behavior policy mixtures used to train the respective offline RL models: * (VAD($\sigma=0.2$) + VAD($\sigma=0.4$)) and $\dag$ (VAD($\sigma=0.2$) + Random). \textbf{Bold} = best, \underline{underline} = 2nd best.}
\label{tab:main_results}
\resizebox{\textwidth}{!}{
\begin{tabular}{@{}lc | cccc | c | cc@{}}
\toprule
 & & \multicolumn{4}{c|}{General Driving} & Safety-Critical\,  \\ 
Method & Type & CR (\%) $\downarrow$ & RC (\%) $\uparrow$ & Long. Jerk (m/s$^3$) $\downarrow$ & Lat. Jerk (m/s$^3$) $\downarrow$ & CR (\%) $\downarrow$ &  SRC $\uparrow$ & JSR $\uparrow$  \\
\midrule
UniAD~\cite{hu2023planning} & IL & 61.3 & \underline{57.4} & \underline{0.33} & \underline{0.21} & 83.0 & 9.8 & 6.6 \\
VAD~\cite{jiang2023vad} & IL & 67.2 & 55.8 & \textbf{0.28} & \textbf{0.19} & 83.1 & 11.4 & 9.4 \\
SparseDrive~\cite{sun2025sparsedrive} & IL & 81.8 & 54.1 & 0.60 & 0.33 & 90.6 & 5.1 & 1.7 \\
VADv2~\cite{chen2024vadv2} & IL & 73.0 & 34.1 & 0.45 & 0.28 & 65.9 & 11.6 & 9.2 \\
VADv2 (w/ Expert BC) & IL & 66.4 & 31.8 & 0.84 & 0.28 & 38.2 & \underline{19.6} & 20.8 \\
VADv2 (w/ std. BC) & Offline RL & 69.3 & 16.0 & 1.17 & 0.38 & \textbf{17.9} & 13.1 & \underline{25.2} \\
VADv2* (w/ Expert BC) & Offline RL & \underline{51.1} & 52.8 & 0.75 & 0.27 & \underline{29.9} & \textbf{37.0} & \textbf{34.3} \\
VADv2$\dag$ (w/ Expert BC) & Offline RL & \textbf{35.8} & \textbf{72.0} & 0.47 & 0.23 & 74.7 & 18.2 & 16.3 \\
\bottomrule
\end{tabular}%
}
\end{table*}

\subsection{Implementation Details}
Our model implementation is based on the VADv2. The discrete action space $\mathcal{A}$ consists of $K=4096$ action prototypes, which were generated by applying k-means clustering to the expert trajectories in the nuScenes dataset.
We initialize the network weights from a model pre-trained using standard IL on the nuScenes expert dataset. The model is then fine-tuned using our offline RL approach with 114K iterations. We use the AdamW optimizer with a base learning rate of $3 \times 10^{-5}$ and a weight decay of 0.01. The learning rate is annealed using a cosine scheduler. We use a batch size of 8. All experiments are conducted using NVIDIA H200 GPUs.
For the actor-critic algorithm, the discount factor $\gamma$ is set to 0.9. The target networks are updated using an EMA with a coefficient of $1 \times 10^{-4}$. Further details on training stability are provided in Appendix~\ref{sec:appendix_training_stability}. For our main results, the pseudo-expert BC weight $\alpha$ is set to 0.1. The reward function weights are $w_\text{imitation}=0.1$ and $w_\text{event}=1.0$. The terminal event penalties are set to $C_\text{event} = C_\text{collision} = C_\text{off-road} = C_\text{off-route} = -10.0$. 

\subsection{Baselines}
We compare our proposed offline RL method against several state-of-the-art E2E autonomous driving models (UniAD, VAD, SparseDrive, VADv2) trained with IL, as well as key ablations of our own method. VADv2 (w/ Expert BC) is an IL baseline trained using only our pseudo-expert behavior cloning loss (without the RL objective) to isolate the contribution of the expert regularization. VADv2 (w/ std. BC) is an offline RL variant of our method. Instead of using our pseudo-expert regularization, it is trained with a standard BC loss that clones the behavior policy $\pi_\beta$ from the dataset $\mathcal{D}$. This baseline measures the impact of cloning a potentially suboptimal policy.

\subsection{Main Results}
We present our main findings in Table~\ref{tab:main_results}, comparing our offline RL approach against the IL baselines.
Our proposed offline RL method, VADv2* (trained on the VAD($\sigma=0.2$) + VAD($\sigma=0.4$) dataset mix), demonstrates a significantly improved balance of safety and driving efficiency compared to all IL baselines. Critically, it achieves the second lowest CR on the challenging Safety-Critical benchmark at 29.9\%. This is a substantial improvement over the best-performing IL baseline (VADv2 w/ Expert BC), which only achieves a 38.2\% CR.
While some IL methods like UniAD achieve a higher RC in general driving (57.4\%), they perform very poorly on the safety-critical suite (83.0\% CR). Our method maintains a strong RC of 52.8\% while excelling in safety. This superior trade-off is captured by the unified metrics: our VADv2* method achieves the highest SRC at 37.0 and the highest JSR at 34.3, outperforming all other methods.

The comparisons with other offline RL variants highlight our design choices. The VADv2 (w/ std. BC) model, which uses standard BC instead of our pseudo-expert regularization, becomes overly conservative, resulting in a catastrophic drop in route completion (16.0\% RC). Conversely, the VADv2$\dag$ model, trained with a Random policy, learns to complete routes (72.0\% RC) but fails to learn safe behavior (74.7\% CR in safety scenarios). This confirms that both the pseudo-expert regularization and a well-composed behavior policy dataset are essential.

We note that the IL baselines achieve lower jerk values. This is an expected outcome, as our reward function did not include an explicit penalty for jerk to optimize ride comfort.

\subsection{Ablation Studies}
\label{sec:ablation}
To further understand the factors contributing to our model's performance, we conduct targeted ablations on the BC regularization weight, reward components, and behavior policy composition.

\begin{table}[h]
\centering
\caption{Ablation on the pseudo-expert BC regularization weight ($\alpha$). The behavior policy mixture was VAD($\sigma=0.2$) + VAD($\sigma=0.4$), and reward parameters were fixed at $w_\text{imitation}=0.1$ and $C_\text{event}=-10$.}
\label{tab:ablation_bc_weight}
\resizebox{0.43\textwidth}{!}{
\begin{tabular}{@{}lcccc@{}}
\toprule
$\alpha$ & \begin{tabular}[c]{@{}c@{}}RC (\%) $\uparrow$\\ (General)\end{tabular} & \begin{tabular}[c]{@{}c@{}}CR (\%) $\downarrow$\\ (Safety-Critical)\end{tabular} & \begin{tabular}[c]{@{}c@{}}SRC $\uparrow$\\ \end{tabular} & \begin{tabular}[c]{@{}c@{}}Jerk (m/s$^3$) $\downarrow$\\ (General)\end{tabular} \\
\midrule
0.0 & 21.6 & 87.3 & 2.7 & 1.67 \\
0.1 & \underline{52.8} & \underline{29.9} & \textbf{37.0} & 0.51 \\
0.2 & 46.5 & \textbf{28.1} & 33.5 & 0.40 \\
0.4 & \textbf{55.3} & 35.9 & \underline{35.4} & \underline{0.38} \\
1.0 & 45.6 & 48.8 & 23.3 & \textbf{0.34} \\
\bottomrule
\end{tabular}%
}
\end{table}

\noindent
\textbf{Effect of BC Weight ($\alpha$).}
As shown in Table~\ref{tab:ablation_bc_weight}, the pseudo-expert regularization weight $\alpha$ has a significant impact. Removing the BC regularization ($\alpha=0.0$) leads to unstable policy updates and very poor performance (87.3\% CR, 21.6\% RC), underscoring the need for regularization to mitigate OOD action exploitation. Conversely, a very high weight ($\alpha=1.0$) appears to over-constrain the policy, degrading both route completion and safety (48.8\% CR). Optimal performance is found in the range $\alpha \in [0.1, 0.4]$.

\begin{table}[h]
\centering
\caption{Ablation on Reward Components. The behavior policy mixture was VAD($\sigma=0.2$) + VAD($\sigma=0.4$), and $\alpha=0.2$.}
\label{tab:ablation_reward}
\resizebox{0.48\textwidth}{!}{
\begin{tabular}{@{}lccccc@{}}
\toprule
$w_\text{imitation}$ & $C_\text{event}$ & \begin{tabular}[c]{@{}c@{}}RC (\%) $\uparrow$\\ (General)\end{tabular} & \begin{tabular}[c]{@{}c@{}}CR (\%) $\downarrow$\\ (Safety-Critical)\end{tabular} & \begin{tabular}[c]{@{}c@{}}SRC $\uparrow$\\ \end{tabular} & \begin{tabular}[c]{@{}c@{}}Jerk (m/s$^3$) $\downarrow$\\ (General)\end{tabular} \\
\midrule
0 & -10 & 26.5 & \underline{33.0} & 17.7 & 0.53 \\
0.1 & -5 & \underline{43.7} & 46.2  & \underline{23.5} & \textbf{0.34} \\
0.1 & -10 & \textbf{46.5} & \textbf{28.1} & \textbf{33.5} & \underline{0.40} \\
0.1 & -20 & 34.5 & 37.2 & 21.7 & 0.45 \\
\bottomrule
\end{tabular}%
}
\end{table}

\noindent
\textbf{Effect of Reward Components.}
Table~\ref{tab:ablation_reward} analyzes reward components. Removing the dense expert imitation reward ($w_\text{imitation}=0$) severely degrades performance, causing RC to drop from 46.5\% to 26.5\%. This demonstrates that the dense imitation signal is critical for guiding the policy to learn efficient driving. The magnitude of the terminal event penalty ($C_\text{event}$) is also crucial. A weak penalty ($C_\text{event}=-5$) is insufficient to prevent collisions, resulting in a high 46.2\% CR in safety scenarios. Conversely, an overly harsh penalty ($C_\text{event}=-20$) makes the policy too conservative, hurting its ability to complete routes (34.5\% RC) without a corresponding improvement in safety (37.2\% CR). The combination of $w_\text{imitation}=0.1$ and $C_\text{event}=-10$ provides the best trade-off, achieving the highest SC score.

\begin{table}[h]
\centering
\caption{Ablation on Behavior Policy. Reward parameters were fixed at $w_\text{imitation}=0.1$ and $C_\text{event}=-10$, and $\alpha=0.2$.}
\label{tab:ablation_data}
\resizebox{0.48\textwidth}{!}{
\begin{tabular}{@{}lcccc@{}}
\toprule
Behavior Policy & \begin{tabular}[c]{@{}c@{}}RC (\%) $\uparrow$\\ (General)\end{tabular} & \begin{tabular}[c]{@{}c@{}}CR (\%) $\downarrow$\\ (Safety-Critical)\end{tabular} & \begin{tabular}[c]{@{}c@{}}SRC $\uparrow$\\ \end{tabular} & \begin{tabular}[c]{@{}c@{}}Jerk (m/s$^3$) $\downarrow$\\ (General)\end{tabular} \\
\midrule
VAD($\sigma=0.2$) & 40.7 & \underline{45.0} & \underline{22.4} & 0.49 \\
VAD($\sigma=0.4$) & \underline{70.9} & 69.4 & 21.7 & 0.42 \\
Random & 67.0 & 74.2 & 17.3 & 0.72\\
VAD($\sigma=0.2$) + Random & \textbf{72.0} & 74.7 & 18.2 & \underline{0.35} \\
VAD($\sigma=0.2$) + VAD($\sigma=0.4$) & 46.5 & \textbf{28.1} & \textbf{33.5} & 0.40 \\
VADv2($\sigma=0.2$) & 56.5 & 83.1 & 9.5 & \textbf{0.33} \\
\bottomrule
\end{tabular}%
}
\end{table}

\noindent
\textbf{Effect of Behavior Policy Composition.}
Finally, Table~\ref{tab:ablation_data} explores the influence of the offline dataset's composition. Using highly stochastic policies like VAD($\sigma=0.4$) or a Random policy leads to agents that complete routes but are extremely unsafe (69.4\% and 74.2\% CR, respectively). The VADv2($\sigma=0.2$) policy led to the worst overall performance (9.5 SC). Our main results (Table~\ref{tab:main_results}) showed that mixing a good policy with Random data (VADv2$\dag$) was also ineffective. This ablation confirms that finding a good mixture is key. The combination of VAD($\sigma=0.2$) and VAD($\sigma=0.4$) yields the best results by far, achieving the lowest safety-critical CR (28.1\%) and the highest SC score (33.5). This suggests an optimal dataset contains a blend of near-optimal driving and meaningful, recoverable deviations. Further analysis is found in Sec.~\ref{sec:behavior_policy_and_learned_strategy}.

\subsection{Behavior Policy and Learned Strategy}
\label{sec:behavior_policy_and_learned_strategy}

\begin{figure}
  \centering
  \begin{subfigure}{0.99\linewidth}
    \includegraphics[width=0.99\linewidth]{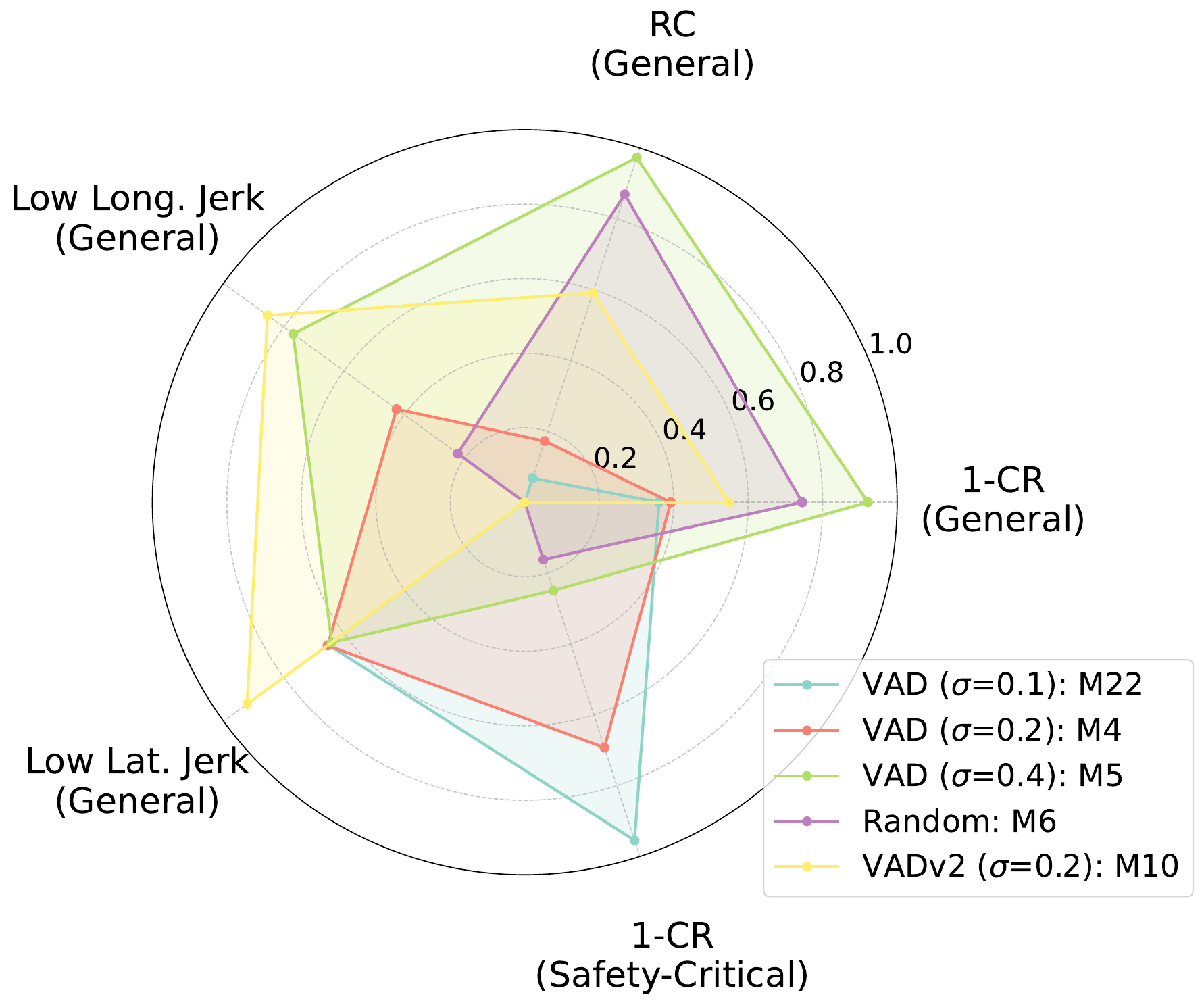}
    \caption{}
    \label{fig:policy_character_a}
  \end{subfigure} \\
  \begin{subfigure}{0.98\linewidth}
    \includegraphics[width=0.98\linewidth]{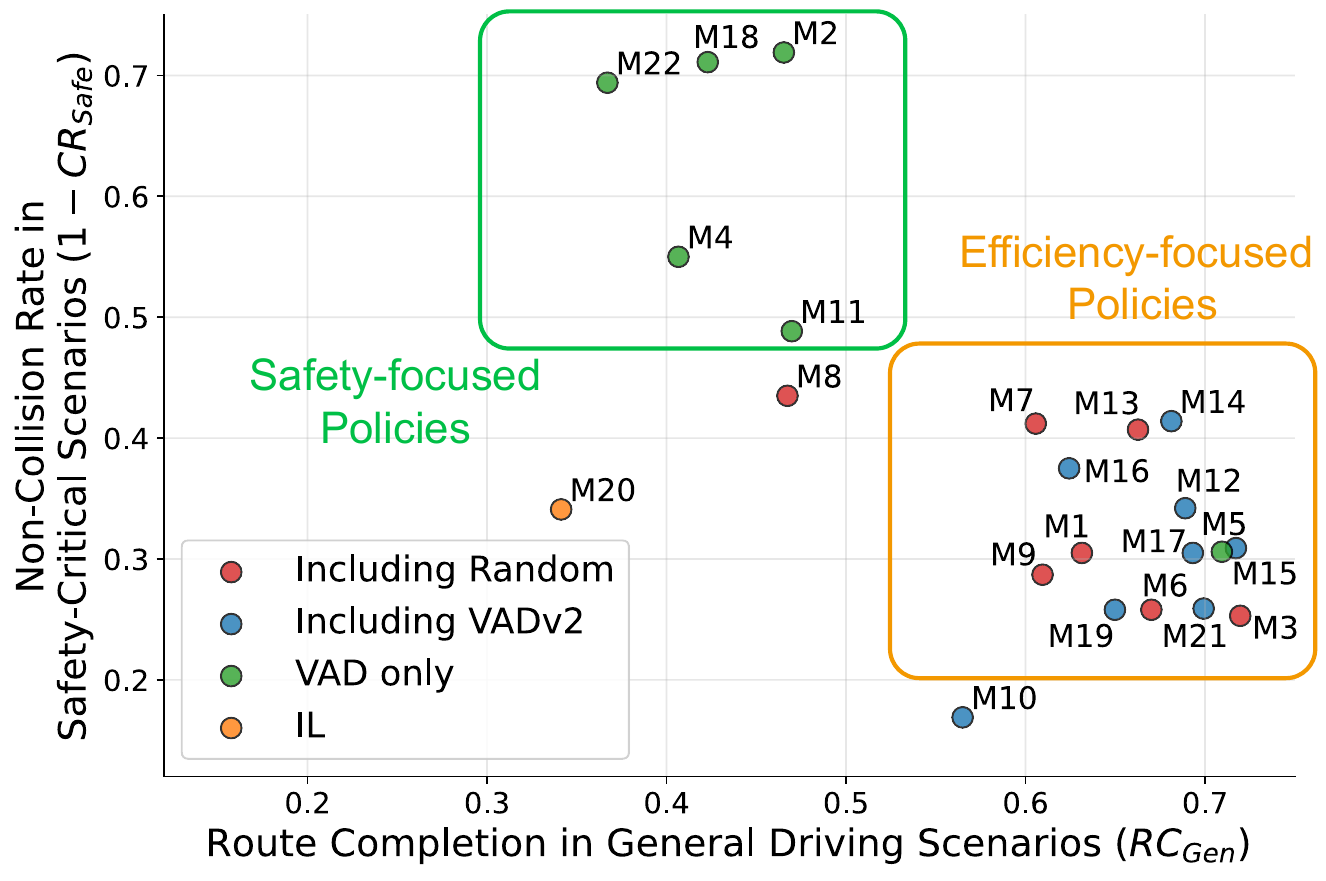}
    \caption{}
    \label{fig:policy_character_b}
  \end{subfigure}
  \caption{Influence of behavior policy on learned strategy. Model IDs (e.g., M4) are consistent across both subfigures. (a) Radar chart comparing policy `personalities' across metrics. All values are normalized with min and max values among all models. (b) Scatter plot showing the trade-off between general driving efficiency ($\text{RC}_\text{Gen}$) and safety-critical non-collision rate ($1-\text{CR}_\text{Safe}$).}
  \label{fig:policy_character}
\end{figure}

\begin{figure*}[htbp]
  \centering
  \begin{subfigure}{0.48\linewidth}
    \includegraphics[width=0.98\linewidth]{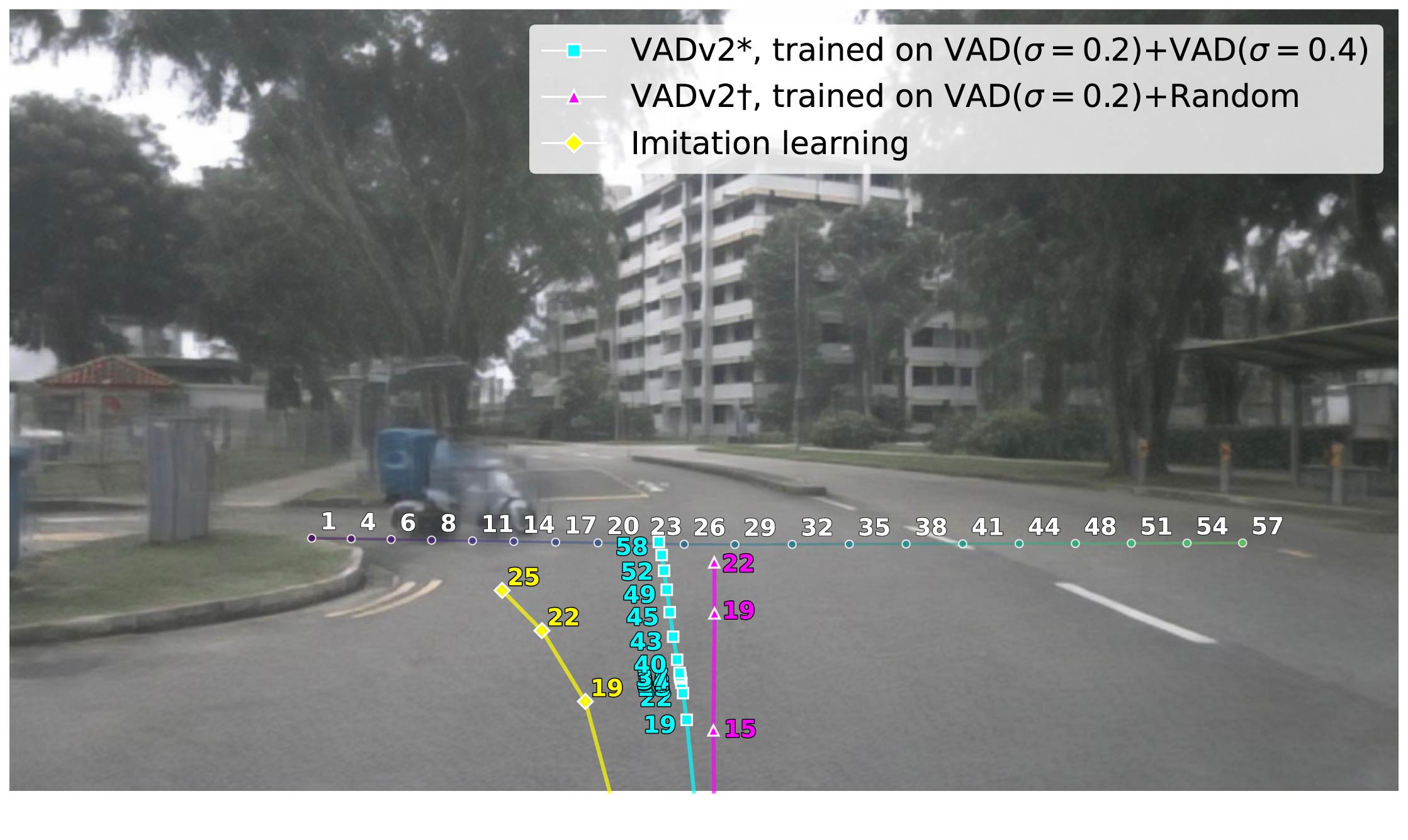}
    \caption{}
    \label{fig:qualitative_results_a}
  \end{subfigure}
  \hfill
  \begin{subfigure}{0.48\linewidth}
    \includegraphics[width=0.98\linewidth]{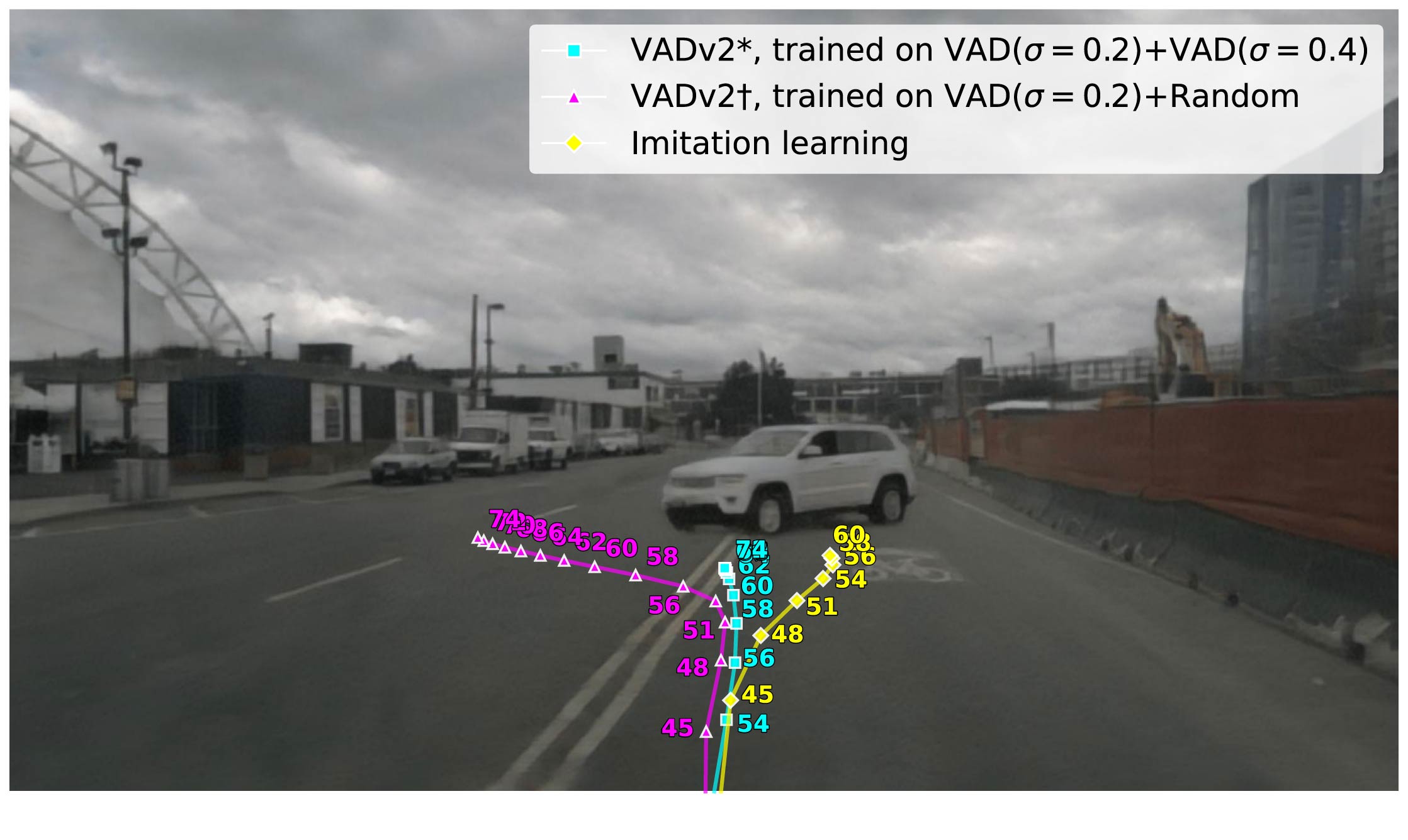}
    \caption{}
    \label{fig:qualitative_results_b}
  \end{subfigure}
  \caption{Qualitative results in two safety-critical NeuroNCAP scenarios. We compare the trajectories of our offline RL model (VADv2*, trained on VAD($\sigma=0.2$) + VAD($\sigma=0.4$) dataset) against a baseline IL model and an RL ablation (VADv2†, trained on VAD($\sigma=0.2$) + Random dataset). The numbers assigned for each waypoint indicate the frame index. (a) A scooter cuts into the ego-vehicle's path. (b) A stationary vehicle obstructs the lane.}
  \label{fig:qualitative_results}
\end{figure*}

We analyze the driving strategies (``personalities'') learned from different behavior datasets, finding that the dataset composition strongly influences the policy's position on the safety-efficiency trade-off.
The full details of the policy mixtures and experimental conditions used for this analysis are provided in Appendix~\ref{sec:appendix_behavior_policy}. 
As shown in Fig.~\ref{fig:policy_character_b}, there is an inverse relationship between general route completion ($\text{RC}_\text{Gen}$) and safety-critical non-collision rates ($1-\text{CR}_\text{Safe}$). We observe that higher data randomness (e.g., more noise, random policy) or the inclusion of the VADv2 policy correlates with an efficiency-focused policy (high $\text{RC}_\text{Gen}$, high $\text{CR}_\text{Safe}$). Conversely, lower randomness (e.g., monotonous behavior from VAD-only data) yields a safety-focused policy (low $\text{RC}_\text{Gen}$, low $\text{CR}_\text{Safe}$).
The VADv2 policy itself appears to introduce a form of exploratory behavior; it tends to sway from side to side, which is supported by its higher lateral jerk compared to VAD (Table~\ref{tab:main_results}). We hypothesize this inherent instability serves as a substitute for explicit randomness. Fig.~\ref{fig:policy_character_a} details this trend: efficiency-focused policies (M5, M6, M10) spread upward in the radar chart, while safety-focused policies (M4, M22) spread downward.

We hypothesize this stems from how the Q-function's value estimates are shaped by the data distribution.
\textbf{High-randomness data} includes a wide-range of actions that includes many implausible ones (e.g., swerving on a straight road). In this context, the Q-value for the `proceed straight' action is learned to be relatively high, as it is clearly superior to these random alternatives. The resulting policy may then favor this high-value `go' action, leading to a strategy that performs well in general scenarios but fails in critical ones where stopping is required.
\textbf{Low-randomness data} is monotonous, reflecting the base IL policy's consistent (and suboptimal) behavior. This dataset repeatedly associates these common actions with collisions. The Q-function thus learns pessimistic Q-values for this specific action set, resulting in a more cautious, safety-focused policy that favors stopping to avoid these known failure modes.

\subsection{Qualitative Results}

Figure~\ref{fig:qualitative_results} visualizes closed-loop trajectories in two safety-critical scenarios from NeuroNCAP. These examples compare the behavior of our primary offline RL model (VADv2*, trained on the VAD($\sigma=0.2$) + VAD($\sigma=0.4$) dataset), an IL baseline, and the offline RL ablation (VADv2†, trained on the VAD($\sigma=0.2$) + Random dataset).

In Fig.~\ref{fig:qualitative_results_a}, an adversarial scooter unexpectedly cuts into the ego-vehicle's path. Both the IL baseline and the VADv2† model fail to react appropriately, proceeding on their path and resulting in a collision. In contrast, the VADv2* model correctly identifies the hazard, executes an avoidance maneuver: it stops before the intersection, allows the scooter to pass, and then resumes driving.
Figure~\ref{fig:qualitative_results_b} illustrates a scenario with a stationary vehicle blocking the ego-vehicle's lane. The IL baseline proceeds without changing course and collides with the obstacle. This scene highlights two distinct avoidance strategies learned by the RL agents. The VADv2† model, an efficient-focused policy (Fig~\ref{fig:policy_character_b}), adopts a swerving maneuver. In contrast, the VADv2* model, a safety-focused policy, employs a more cautious policy, opting to come to a complete stop behind the obstructing vehicle. Additional qualitative results are available in Appendix~\ref{sec:appendix_qualitative}.

\section{Conclusion}
\label{sec:conclusion}

We proposed a camera-only, E2E offline RL framework for autonomous driving, learning from a fixed dataset to avoid the high costs of online RL and the limitations of IL. We experimentally confirmed that using the pseudo-expert regularization technique helps stabilize training and suppress imitating unsafe behaviors, unlike standard BC. Closed-loop evaluations demonstrated that our method substantially improves safety and efficiency compared to IL baselines, particularly in critical scenarios. Ablation studies confirmed the importance of our regularization and the significant influence of behavior policy composition. 

A critical yet underexplored question in offline RL concerns the principled selection of behavior policies for dataset collection. In this study, we investigate this question empirically by evaluating various behavior policies and their mixtures. While we do not yet establish comprehensive criteria for behavior policy selection, our results reveal that the characteristics of the behavior policy significantly influence the properties of the trained target policy in the offline setting.
Additionally, we incorporate BC alongside RL objectives to stabilize training, leveraging empirically defined pseudo-expert GT trajectories. Future work includes the principled selection of behavior policies and the systematic construction of pseudo-expert trajectories.

{
    \small
    \bibliographystyle{ieeenat_fullname}
    \bibliography{main}

@String(CVPR= {IEEE Conf. Comput. Vis. Pattern Recog.})

@String(ICCV= {Int. Conf. Comput. Vis.})

@String(ECCV= {Eur. Conf. Comput. Vis.})

@String(ICLR = {Int. Conf. Learn. Represent.})

@String(CVPR  = {CVPR})

@String(ICCV  = {ICCV})

@String(ECCV  = {ECCV})

@String(ICLR  = {ICLR})

@inproceedings{li2024think2drive,
  title={Think2drive: Efficient reinforcement learning by thinking with latent world model for autonomous driving (in carla-v2)},
  author={Li, Qifeng and Jia, Xiaosong and Wang, Shaobo and Yan, Junchi},
  booktitle={ECCV},
  year={2024},
}

@inproceedings{gao2025rad,
  title={Rad: Training an end-to-end driving policy via large-scale 3dgs-based reinforcement learning},
  author={Gao, Hao and Chen, Shaoyu and Jiang, Bo and Liao, Bencheng and Shi, Yiang and Guo, Xiaoyang and Pu, Yuechuan and Yin, Haoran and Li, Xiangyu and Zhang, Xinbang and others},
  booktitle={NeurIPS},
  year={2025}
}

@article{li2025recogdrive,
  title={ReCogDrive: A Reinforced Cognitive Framework for End-to-End Autonomous Driving},
  author={Li, Yongkang and Xiong, Kaixin and Guo, Xiangyu and Li, Fang and Yan, Sixu and Xu, Gangwei and Zhou, Lijun and Chen, Long and Sun, Haiyang and Wang, Bing and others},
  journal={arXiv preprint arXiv:2506.08052},
  year={2025}
}

@inproceedings{dosovitskiy2017carla,
  title={CARLA: An open urban driving simulator},
  author={Dosovitskiy, Alexey and Ros, German and Codevilla, Felipe and Lopez, Antonio and Koltun, Vladlen},
  booktitle={CoRL},
  year={2017},
}

@inproceedings{toromanoff2020end,
  title={End-to-end model-free reinforcement learning for urban driving using implicit affordances},
  author={Toromanoff, Marin and Wirbel, Emilie and Moutarde, Fabien},
  booktitle={CVPR},
  year={2020}
}

@inproceedings{liang2018cirl,
  title={Cirl: Controllable imitative reinforcement learning for vision-based self-driving},
  author={Liang, Xiaodan and Wang, Tairui and Yang, Luona and Xing, Eric},
  booktitle={ECCV},
  year={2018}
}

@inproceedings{lu2023imitation,
  title={Imitation is not enough: Robustifying imitation with reinforcement learning for challenging driving scenarios},
  author={Lu, Yiren and Fu, Justin and Tucker, George and Pan, Xinlei and Bronstein, Eli and Roelofs, Rebecca and Sapp, Benjamin and White, Brandyn and Faust, Aleksandra and Whiteson, Shimon and others},
  booktitle={IROS},
  year={2023},
}

@inproceedings{hu2024solving,
  title={Solving motion planning tasks with a scalable generative model},
  author={Hu, Yihan and Chai, Siqi and Yang, Zhening and Qian, Jingyu and Li, Kun and Shao, Wenxin and Zhang, Haichao and Xu, Wei and Liu, Qiang},
  booktitle={ECCV},
  year={2024},
}

@inproceedings{chen2021learning,
  title={Learning to drive from a world on rails},
  author={Chen, Dian and Koltun, Vladlen and Kr{\"a}henb{\"u}hl, Philipp},
  booktitle={ICCV},
  year={2021}
}

@article{chen2021interpretable,
  title={Interpretable end-to-end urban autonomous driving with latent deep reinforcement learning},
  author={Chen, Jianyu and Li, Shengbo Eben and Tomizuka, Masayoshi},
  journal={IEEE ITS},
  year={2021},
}

@inproceedings{zhang2021end,
  title={End-to-end urban driving by imitating a reinforcement learning coach},
  author={Zhang, Zhejun and Liniger, Alexander and Dai, Dengxin and Yu, Fisher and Van Gool, Luc},
  booktitle={ICCV},
  year={2021}
}

@inproceedings{ye2025lord,
  title={Lord: Large models based opposite reward design for autonomous driving},
  author={Ye, Xin and Tao, Feng and Mallik, Abhirup and Yaman, Burhaneddin and Ren, Liu},
  booktitle={WACV},
  year={2025},
}

@article{huang2025vlm,
  title={Vlm-rl: A unified vision language models and reinforcement learning framework for safe autonomous driving},
  author={Huang, Zilin and Sheng, Zihao and Qu, Yansong and You, Junwei and Chen, Sikai},
  journal={Transp. Res. Part C},
  year={2025},
}

@article{shao2024deepseekmath,
  title={Deepseekmath: Pushing the limits of mathematical reasoning in open language models},
  author={Shao, Zhihong and Wang, Peiyi and Zhu, Qihao and Xu, Runxin and Song, Junxiao and Bi, Xiao and Zhang, Haowei and Zhang, Mingchuan and Li, YK and Wu, Yang and others},
  journal={arXiv preprint arXiv:2402.03300},
  year={2024}
}

@article{dauner2024navsim,
  title={Navsim: Data-driven non-reactive autonomous vehicle simulation and benchmarking},
  author={Dauner, Daniel and Hallgarten, Marcel and Li, Tianyu and Weng, Xinshuo and Huang, Zhiyu and Yang, Zetong and Li, Hongyang and Gilitschenski, Igor and Ivanovic, Boris and Pavone, Marco and others},
  journal={NeurIPS},
  year={2024}
}

@article{schulman2017proximal,
  title={Proximal policy optimization algorithms},
  author={Schulman, John and Wolski, Filip and Dhariwal, Prafulla and Radford, Alec and Klimov, Oleg},
  journal={arXiv preprint arXiv:1707.06347},
  year={2017}
}

@article{nair2020awac,
  title={Awac: Accelerating online reinforcement learning with offline datasets},
  author={Nair, Ashvin and Gupta, Abhishek and Dalal, Murtaza and Levine, Sergey},
  journal={arXiv preprint arXiv:2006.09359},
  year={2020}
}

@article{fujimoto2021minimalist,
  title={A minimalist approach to offline reinforcement learning},
  author={Fujimoto, Scott and Gu, Shixiang Shane},
  journal={NeurIPS},
  year={2021}
}

@article{tarasov2023revisiting,
  title={Revisiting the minimalist approach to offline reinforcement learning},
  author={Tarasov, Denis and Kurenkov, Vladislav and Nikulin, Alexander and Kolesnikov, Sergey},
  journal={NeurIPS},
  year={2023}
}

@inproceedings{wang2022diffusion,
  title={Diffusion policies as an expressive policy class for offline reinforcement learning},
  author={Wang, Zhendong and Hunt, Jonathan J and Zhou, Mingyuan},
  booktitle={ICLR},
  year={2023}
}

@inproceedings{park2025flow,
  title={Flow q-learning},
  author={Park, Seohong and Li, Qiyang and Levine, Sergey},
  booktitle={ICML},
  year={2025}
}

@article{kumar2020conservative,
  title={Conservative q-learning for offline reinforcement learning},
  author={Kumar, Aviral and Zhou, Aurick and Tucker, George and Levine, Sergey},
  journal={NeurIPS},
  year={2020}
}

@inproceedings{kostrikov2021offline,
  title={Offline reinforcement learning with implicit q-learning},
  author={Kostrikov, Ilya and Nair, Ashvin and Levine, Sergey},
  booktitle={ICLR},
  year={2022}
}

@inproceedings{xu2023offline,
  title={Offline rl with no ood actions: In-sample learning via implicit value regularization},
  author={Xu, Haoran and Jiang, Li and Li, Jianxiong and Yang, Zhuoran and Wang, Zhaoran and Chan, Victor Wai Kin and Zhan, Xianyuan},
  booktitle={ICLR},
  year={2023}
}

@inproceedings{garg2023extreme,
  title={Extreme q-learning: Maxent rl without entropy},
  author={Garg, Divyansh and Hejna, Joey and Geist, Matthieu and Ermon, Stefano},
  booktitle={ICLR},
  year={2023}
}

@article{hansen2023idql,
  title={Idql: Implicit q-learning as an actor-critic method with diffusion policies},
  author={Hansen-Estruch, Philippe and Kostrikov, Ilya and Janner, Michael and Kuba, Jakub Grudzien and Levine, Sergey},
  journal={arXiv preprint arXiv:2304.10573},
  year={2023}
}

@article{yu2020mopo,
  title={Mopo: Model-based offline policy optimization},
  author={Yu, Tianhe and Thomas, Garrett and Yu, Lantao and Ermon, Stefano and Zou, James Y and Levine, Sergey and Finn, Chelsea and Ma, Tengyu},
  journal={NeurIPS},
  year={2020}
}

@article{kidambi2020morel,
  title={Morel: Model-based offline reinforcement learning},
  author={Kidambi, Rahul and Rajeswaran, Aravind and Netrapalli, Praneeth and Joachims, Thorsten},
  journal={NeurIPS},
  year={2020}
}

@article{an2021uncertainty,
  title={Uncertainty-based offline reinforcement learning with diversified q-ensemble},
  author={An, Gaon and Moon, Seungyong and Kim, Jang-Hyun and Song, Hyun Oh},
  journal={NeurIPS},
  year={2021}
}

@article{yu2021combo,
  title={Combo: Conservative offline model-based policy optimization},
  author={Yu, Tianhe and Kumar, Aviral and Rafailov, Rafael and Rajeswaran, Aravind and Levine, Sergey and Finn, Chelsea},
  journal={NeurIPS},
  year={2021}
}

@article{chen2021decision,
  title={Decision transformer: Reinforcement learning via sequence modeling},
  author={Chen, Lili and Lu, Kevin and Rajeswaran, Aravind and Lee, Kimin and Grover, Aditya and Laskin, Misha and Abbeel, Pieter and Srinivas, Aravind and Mordatch, Igor},
  journal={NeurIPS},
  year={2021}
}

@article{janner2021offline,
  title={Offline reinforcement learning as one big sequence modeling problem},
  author={Janner, Michael and Li, Qiyang and Levine, Sergey},
  journal={NeurIPS},
  year={2021}
}

@inproceedings{janner2022planning,
  title={Planning with diffusion for flexible behavior synthesis},
  author={Janner, Michael and Du, Yilun and Tenenbaum, Joshua B and Levine, Sergey},
  booktitle={ICML},
  year={2022}
}

@inproceedings{hu2023planning,
  title={Planning-oriented autonomous driving},
  author={Hu, Yihan and Yang, Jiazhi and Chen, Li and Li, Keyu and Sima, Chonghao and Zhu, Xizhou and Chai, Siqi and Du, Senyao and Lin, Tianwei and Wang, Wenhai and others},
  booktitle={CVPR},
  year={2023}
}

@inproceedings{jiang2023vad,
  title={Vad: Vectorized scene representation for efficient autonomous driving},
  author={Jiang, Bo and Chen, Shaoyu and Xu, Qing and Liao, Bencheng and Chen, Jiajie and Zhou, Helong and Zhang, Qian and Liu, Wenyu and Huang, Chang and Wang, Xinggang},
  booktitle={ICCV},
  year={2023}
}

@article{jiang2024senna,
  title={Senna: Bridging large vision-language models and end-to-end autonomous driving},
  author={Jiang, Bo and Chen, Shaoyu and Liao, Bencheng and Zhang, Xingyu and Yin, Wei and Zhang, Qian and Huang, Chang and Liu, Wenyu and Wang, Xinggang},
  journal={arXiv preprint arXiv:2410.22313},
  year={2024}
}

@inproceedings{renz2025simlingo,
  title={Simlingo: Vision-only closed-loop autonomous driving with language-action alignment},
  author={Renz, Katrin and Chen, Long and Arani, Elahe and Sinavski, Oleg},
  booktitle={CVPR},
  year={2025}
}

@inproceedings{sun2025sparsedrive,
  title={Sparsedrive: End-to-end autonomous driving via sparse scene representation},
  author={Sun, Wenchao and Lin, Xuewu and Shi, Yining and Zhang, Chuang and Wu, Haoran and Zheng, Sifa},
  booktitle={ICRA},
  year={2025},
}

@inproceedings{weng2024drive,
  title={Para-drive: Parallelized architecture for real-time autonomous driving},
  author={Weng, Xinshuo and Ivanovic, Boris and Wang, Yan and Wang, Yue and Pavone, Marco},
  booktitle={CVPR},
  year={2024}
}

@article{jia2025drivetransformer,
  title={Drivetransformer: Unified transformer for scalable end-to-end autonomous driving},
  author={Jia, Xiaosong and You, Junqi and Zhang, Zhiyuan and Yan, Junchi},
  journal={ICLR},
  year={2025}
}

@inproceedings{liao2025diffusiondrive,
  title={Diffusiondrive: Truncated diffusion model for end-to-end autonomous driving},
  author={Liao, Bencheng and Chen, Shaoyu and Yin, Haoran and Jiang, Bo and Wang, Cheng and Yan, Sixu and Zhang, Xinbang and Li, Xiangyu and Zhang, Ying and Zhang, Qian and others},
  booktitle={CVPR},
  year={2025}
}

@inproceedings{zheng2024genad,
  title={Genad: Generative end-to-end autonomous driving},
  author={Zheng, Wenzhao and Song, Ruiqi and Guo, Xianda and Zhang, Chenming and Chen, Long},
  booktitle={ECCV},
  year={2024},
}

@article{chen2024vadv2,
  title={Vadv2: End-to-end vectorized autonomous driving via probabilistic planning},
  author={Chen, Shaoyu and Jiang, Bo and Gao, Hao and Liao, Bencheng and Xu, Qing and Zhang, Qian and Huang, Chang and Liu, Wenyu and Wang, Xinggang},
  journal={arXiv preprint arXiv:2402.13243},
  year={2024}
}

@article{li2024hydra,
  title={Hydra-mdp: End-to-end multimodal planning with multi-target hydra-distillation},
  author={Li, Zhenxin and Li, Kailin and Wang, Shihao and Lan, Shiyi and Yu, Zhiding and Ji, Yishen and Li, Zhiqi and Zhu, Ziyue and Kautz, Jan and Wu, Zuxuan and others},
  journal={arXiv preprint arXiv:2406.06978},
  year={2024}
}

@inproceedings{li2024bevformer,
  title={BEVFormer: Learning Bird's-Eye-View Representation From LiDAR-Camera Via Spatiotemporal Transformers},
  author={Li, Zhiqi and Wang, Wenhai and Li, Hongyang and Xie, Enze and Sima, Chonghao and Lu, Tong and Yu, Qiao and Dai, Jifeng},
  booktitle={ECCV},
  year={2022},
}

@inproceedings{
zhu2021deformable,
title={Deformable DETR: Deformable Transformers for End-to-End Object Detection},
author={Xizhou Zhu and Weijie Su and Lewei Lu and Bin Li and Xiaogang Wang and Jifeng Dai},
booktitle={ICLR},
year={2021},
}

@article{levine2020offline,
  title={Offline reinforcement learning: Tutorial, review, and perspectives on open problems},
  author={Levine, Sergey and Kumar, Aviral and Tucker, George and Fu, Justin},
  journal={arXiv preprint arXiv:2005.01643},
  year={2020}
}

@inproceedings{degris2012off,
  title={Off-policy actor-critic},
  author={Degris, Thomas and White, Martha and Sutton, Richard S},
  booktitle={ICML},
  year={2012}
}

@inproceedings{ljungbergh2024neuroncap,
  title={Neuroncap: Photorealistic closed-loop safety testing for autonomous driving},
  author={Ljungbergh, William and Tonderski, Adam and Johnander, Joakim and Caesar, Holger and {\AA}str{\"o}m, Kalle and Felsberg, Michael and Petersson, Christoffer},
  booktitle={ECCV},
  year={2024},
}

@inproceedings{tonderski2024neurad,
  title={Neurad: Neural rendering for autonomous driving},
  author={Tonderski, Adam and Lindstr{\"o}m, Carl and Hess, Georg and Ljungbergh, William and Svensson, Lennart and Petersson, Christoffer},
  booktitle={CVPR},
  year={2024}
}

@inproceedings{caesar2020nuscenes,
  title={nuscenes: A multimodal dataset for autonomous driving},
  author={Caesar, Holger and Bankiti, Varun and Lang, Alex H and Vora, Sourabh and Liong, Venice Erin and Xu, Qiang and Krishnan, Anush and Pan, Yu and Baldan, Giancarlo and Beijbom, Oscar},
  booktitle={CVPR},
  year={2020}
}

@article{de2019causal,
  title={Causal confusion in imitation learning},
  author={De Haan, Pim and Jayaraman, Dinesh and Levine, Sergey},
  journal={Advances in neural information processing systems},
  year={2019}
}

@article{kerbl20233d,
  title={3D Gaussian splatting for real-time radiance field rendering.},
  author={Kerbl, Bernhard and Kopanas, Georgios and Leimk{\"u}hler, Thomas and Drettakis, George},
  journal={ACM Trans. Graph.},
  year={2023}
}

@inproceedings{chen2019top,
  title={Top-k off-policy correction for a REINFORCE recommender system},
  author={Chen, Minmin and Beutel, Alex and Covington, Paul and Jain, Sagar and Belletti, Francois and Chi, Ed H},
  booktitle={WSDM},
  year={2019}
}

@article{jaques2019way,
  title={Way off-policy batch deep reinforcement learning of implicit human preferences in dialog},
  author={Jaques, Natasha and Ghandeharioun, Asma and Shen, Judy Hanwen and Ferguson, Craig and Lapedriza, Agata and Jones, Noah and Gu, Shixiang and Picard, Rosalind},
  journal={NeurIPS workshop},
  year={2019}
}

@article{guo2025deepseek,
  title={Deepseek-r1: Incentivizing reasoning capability in llms via reinforcement learning},
  author={Guo, Daya and Yang, Dejian and Zhang, Haowei and Song, Junxiao and Zhang, Ruoyu and Xu, Runxin and Zhu, Qihao and Ma, Shirong and Wang, Peiyi and Bi, Xiao and others},
  journal={arXiv preprint arXiv:2501.12948},
  year={2025}
}

@article{cusumano2025robust,
  title={Robust autonomy emerges from self-play},
  author={Cusumano-Towner, Marco and Hafner, David and Hertzberg, Alex and Huval, Brody and Petrenko, Aleksei and Vinitsky, Eugene and Wijmans, Erik and Killian, Taylor and Bowers, Stuart and Sener, Ozan and others},
  journal={arXiv preprint arXiv:2502.03349},
  year={2025}
}

@inproceedings{fujimoto2019off,
  title={Off-policy deep reinforcement learning without exploration},
  author={Fujimoto, Scott and Meger, David and Precup, Doina},
  booktitle={ICML},
  year={2019},
}
}


\appendix

\clearpage

\twocolumn[
    \begin{center}
        {\Large \bfseries Pseudo-Expert Regularized Offline RL for End-to-End Autonomous Driving \\ in Photorealistic Closed-Loop Environments} 
    \end{center}
    \begin{center}
        {\Large Supplementary Material} 
    \end{center}
    \vspace{10pt}
]

\section{Detailed Experimental Results for Safety-Efficiency Trade-off}
\label{sec:appendix_behavior_policy}

In this section, we provide the full, detailed results complementing the behavior policy analysis in Section~\ref{sec:behavior_policy_and_learned_strategy} of the main paper.

Table~\ref{tab:appendix_policy_mixtures} lists the complete set of behavior policy compositions evaluated, along with their raw performance metrics on both the General Driving and Safety-Critical benchmarks. The ``Mixing Ratio'' column specifies the data proportions; for example, model M8 was trained on a dataset with a 5:6:1 ratio of VAD ($\sigma=0.2$), VAD ($\sigma=0.4$), and Random policy data, respectively.

\begin{figure*}[h]
  \centering
   \includegraphics[width=0.95\linewidth]{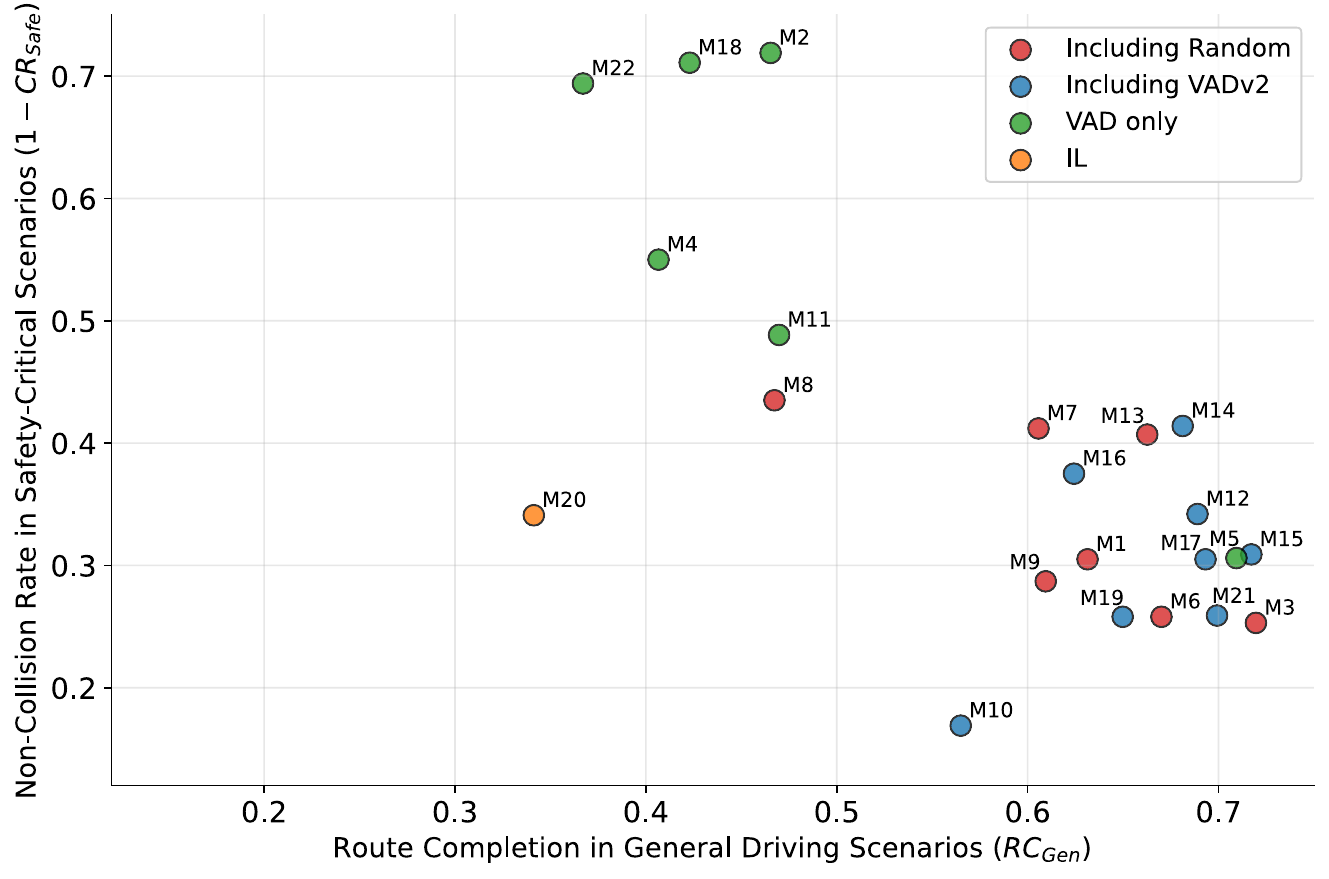}
   \caption{Scatter plot showing the trade-off between route completion in general driving scenarios and non-collision rate in safety-critical scenarios.}
   \label{fig:app_val_vs_safety_scatter}
\end{figure*}

\begin{figure*}[h]
  \centering
   \includegraphics[width=0.95\linewidth]{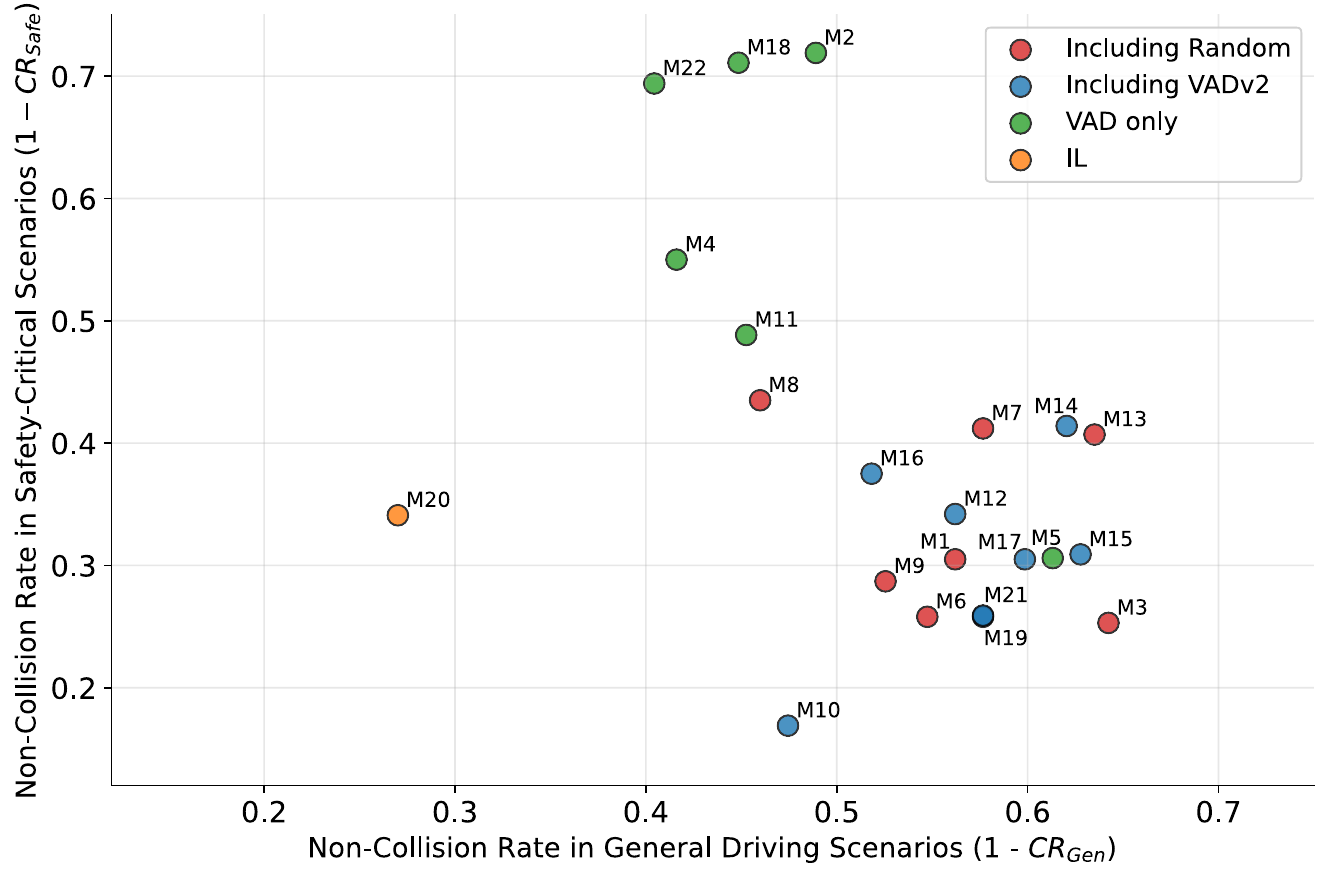}
   \caption{Scatter plot showing the trade-off between non-collision rate in general driving scenarios and non-collision rate in safety-critical scenarios.}
   \label{fig:app_safety_vs_safety_scatter}
\end{figure*}

Additional visualizations are provided to further illustrate the discussed trade-offs. Figure~\ref{fig:app_val_vs_safety_scatter} (which is identical to Figure~\ref{fig:policy_character_b} in the main text) and Figure~\ref{fig:app_safety_vs_safety_scatter} are scatter plots that visualize the relationship between driving efficiency and safety. In these plots, data points are color-coded into four categories based on the training data composition: (1) \textbf{VAD only}: Policies using VAD with varying noise levels; (2) \textbf{Including Random}: Policies trained on VAD data mixed with random policy data; (3) \textbf{Including VADv2}: Policies trained on VAD data mixed with VADv2 data; and (4) \textbf{IL baseline}.

As discussed in Sec.~\ref{sec:behavior_policy_and_learned_strategy}, we observe a clear trend: mixing in VADv2 data or increasing data randomness (higher noise or more random policy data) generally pushes the resulting policy toward the ``efficiency-focused'' region of the graph (bottom-right).

As described in Sec.~\ref{subsec:metrics}, SRC and JSR are defined as the product of the two axes of Figure~\ref{fig:app_val_vs_safety_scatter} and \ref{fig:app_safety_vs_safety_scatter}, respectively. Although the two plots show a similar tendency, they highlight different aspects of performance.
Because both axes of Figure~\ref{fig:app_safety_vs_safety_scatter} are based on non-collision rates, it particularly emphasizes policies that prioritize safety, even at the expense of efficiency. This structure explains why the JSR metric (the product of these two axes) can reward an overly conservative policy. A clear example is the VADv2 (w/ std. BC) model from Table~\ref{tab:main_results}, which achieves a high JSR (25.2\%) precisely because it is extremely safety-focused. In contrast, its modest SRC (13.1\%) score reflects this trade-off, as the SRC metric is designed to measure both efficiency and safety, penalizing the policy for its low route completion.

\begin{figure*}[htbp]
  \centering
  \begin{subfigure}{0.48\linewidth}
    \includegraphics[width=0.98\linewidth]{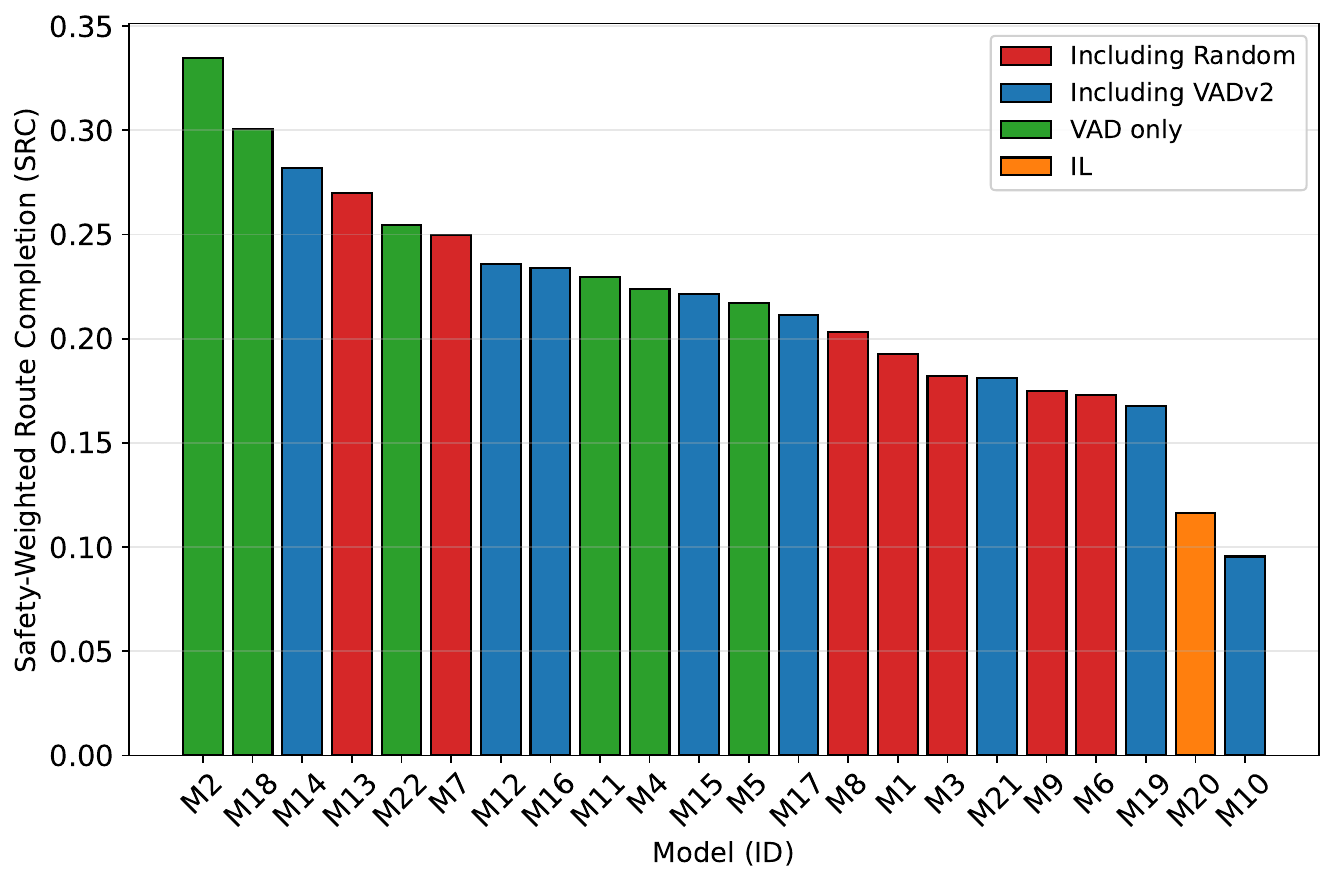}
    \caption{}
    \label{fig:app_src}
  \end{subfigure}
  \hfill
  \begin{subfigure}{0.48\linewidth}
    \includegraphics[width=0.98\linewidth]{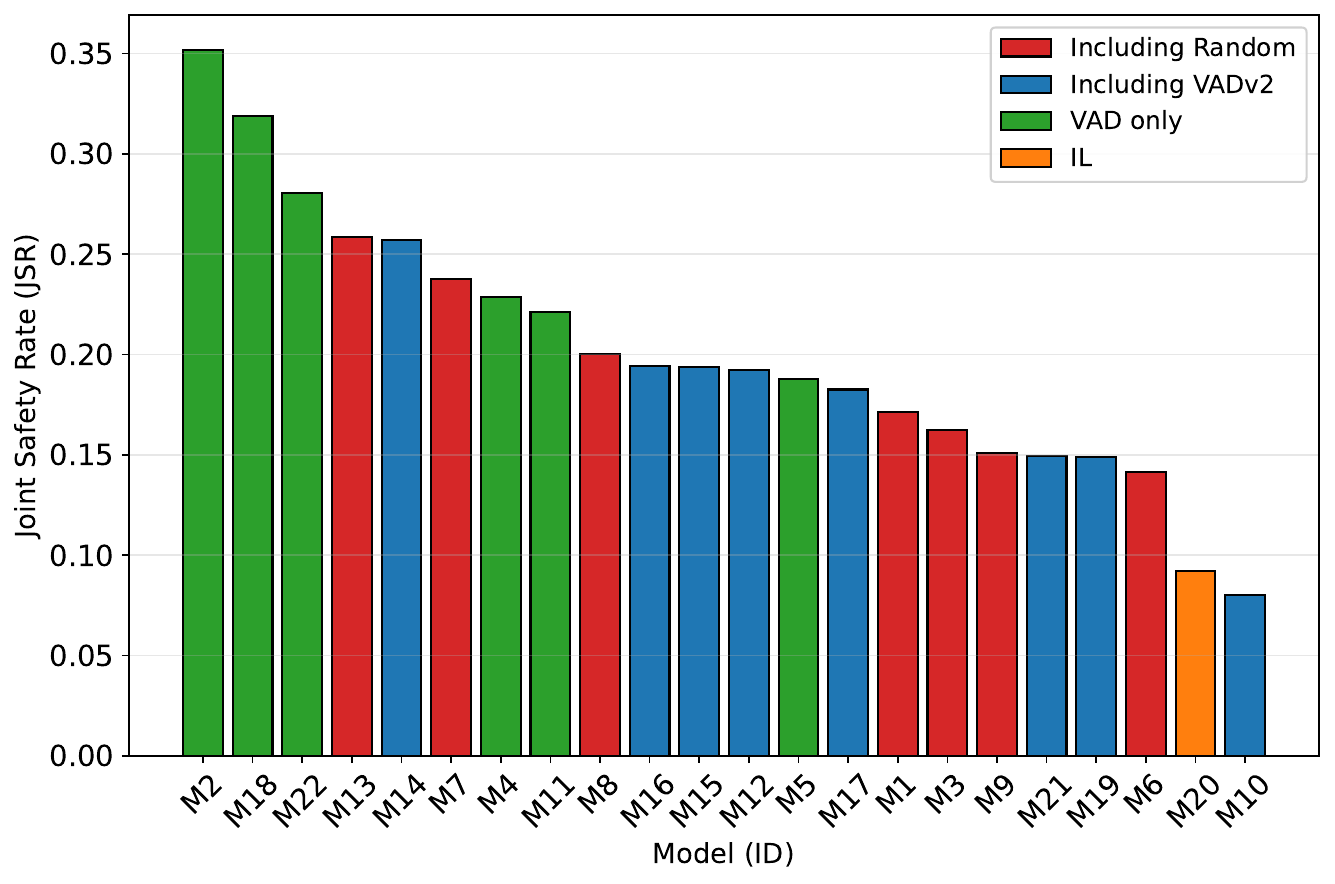}
    \caption{}
    \label{fig:app_jsr}
  \end{subfigure}
  \caption{Comparison of (a) Safety-Weighted Route Completion (SRC) scores and (b) Joint Safery Rate (JSR)  across all evaluated models. The models are sorted by each score in descending order. The Model IDs  (e.g., M2, M18, M10)  correspond to the specific behavior policy compositions detailed in Table~\ref{tab:appendix_policy_mixtures} in Appendix~\ref{sec:appendix_behavior_policy}.}
  \label{fig:src_jsr_bar}
\end{figure*}

Finally, Figure~\ref{fig:src_jsr_bar} presents bar charts ranking all evaluated models by their unified SRC and JSR scores, offering a clear comparison of their balanced performance. These charts illustrate that the VAD only group tends to achieve higher performance than the other groups, and that nearly all offline RL models outperform the IL baseline on these metrics.

\begin{table*}[htbp]
  \centering
  \caption{Dataset compositions for behavior policy analysis (Sec.~\ref{sec:behavior_policy_and_learned_strategy}). Reward parameters were fixed at $w_\text{imitation}=0.1$ and $C_\text{event}=-10$, and $\alpha=0.2$.}
  \label{tab:metrics}
  \footnotesize
  \resizebox{0.98\textwidth}{!}{
  \begin{tabular}{@{} l p{5.5cm} c | c c c c | c | c c c @{}}
    \toprule
    & & &  \multicolumn{4}{c|}{\textbf{General Driving}} & \textbf{Safety-Critical} & &  \\
    \textbf{ID} &
    \textbf{Behavior Policy} &
    \textbf{Mixing Ratio} &
    \textbf{CR $\downarrow$} &
    \textbf{RC $\uparrow$} &
    \textbf{Long. Jerk (m/s$^3$) $\downarrow$} &
    \textbf{Lat. Jerk (m/s$^3$) $\downarrow$} &
    \textbf{CR $\downarrow$} &
    \textbf{SRC $\uparrow$} &
    \textbf{JSR $\uparrow$} \\
    \midrule
    M2 & VAD ($\sigma=0.2$) + VAD ($\sigma=0.4$) & 1:1 & 0.511 & 0.465 & 0.636 & 0.158 & 0.281 & 0.335 & 0.352 \\
    M18 &  VAD ($\sigma=0.1$) + VAD ($\sigma=0.2$)& 1:1 & 0.551 & 0.423 & 0.691 & 0.288 & 0.289 & 0.301 & 0.319 \\
    M14 & VAD ($\sigma=0.1$) + VAD ($\sigma=0.2$) + VAD ($\sigma=0.4$) + VADv2 ($\sigma=0.1$) + VADv2 ($\sigma=0.2$) + VADv2 ($\sigma=0.4$)& 1:1:1:1:1:1  & 0.38 & 0.681 & 0.747 & 0.291 & 0.586 & 0.282 & 0.257 \\
    M13 & VAD ($\sigma=0.1$) + VAD ($\sigma=0.2$) + VAD ($\sigma=0.4$) + VADv2 ($\sigma=0.1$) + VADv2 ($\sigma=0.2$) + Random & 1:1:1:1:1:1 & 0.365 & 0.663 & 0.695 & 0.359 & 0.593 & 0.27 & 0.258 \\
    M22 & VAD ($\sigma=0.1$) & -- & 0.596 & 0.367 & 0.789 & 0.357 & 0.306 & 0.255 & 0.281 \\
    M7 & VAD ($\sigma=0.2$) + VAD ($\sigma=0.4$) + Random & 5:5:2 & 0.423 & 0.606 & 0.457 & 0.248 & 0.588 & 0.25 & 0.238 \\
    M12 & VAD ($\sigma=0.1$) + VAD ($\sigma=0.2$) + VAD ($\sigma=0.4$) + VADv2 ($\sigma=0.2$) & 1:1:1:1 & 0.438 & 0.689 & 0.58 & 0.312 & 0.658 & 0.236 & 0.192 \\
    M16 & VAD ($\sigma=0.2$) + VAD ($\sigma=0.4$) + VADv2 ($\sigma=0.1$) + VADv2 ($\sigma=0.2$) & 1:1:1:1 & 0.482 & 0.624 & 0.454 & 0.257 & 0.625 & 0.234 & 0.194 \\
    M11 & VAD ($\sigma=0.1$) + VAD ($\sigma=0.2$) + VAD ($\sigma=0.4$) & 1:1:1 & 0.547 & 0.47 & 0.55 & 0.217 & 0.512 & 0.229 & 0.221 \\
    M4 & VAD ($\sigma=0.2$) & -- & 0.584 & 0.407 & 0.619 & 0.356 & 0.45 & 0.224 & 0.229 \\
    M15 & VAD ($\sigma=0.1$) + VAD ($\sigma=0.2$) + VAD ($\sigma=0.4$) + VADv2 ($\sigma=0.1$) & 1:1:1:1 & 0.372 & 0.717 & 0.472 & 0.298 & 0.691 & 0.222 & 0.194 \\
    M5 & VAD ($\sigma=0.4$) & -- & 0.387 & 0.709 & 0.482 & 0.363 & 0.694 & 0.217 & 0.188 \\
    M17 & VAD ($\sigma=0.2$) + VAD ($\sigma=0.4$) + VADv2 ($\sigma=0.1$)  & 1:1:1 & 0.401 & 0.693 & 0.437 & 0.243 & 0.695 & 0.211 & 0.183 \\
    M8 & VAD ($\sigma=0.2$) + VAD ($\sigma=0.4$) + Random  & 5:6:1 & 0.54 & 0.467 & 0.389 & 0.2 & 0.565 & 0.203 & 0.2 \\
    M1 & VAD ($\sigma=0.2$) + VAD ($\sigma=0.4$) + Random & 1:1:1 & 0.438 & 0.631 & 0.54 & 0.357 & 0.695 & 0.193 & 0.171 \\
    M3 & VAD ($\sigma=0.2$) + Random & 1:1 & 0.358 & 0.72 & 0.466 & 0.23 & 0.747 & 0.182 & 0.163 \\
    M21 & VAD ($\sigma=0.1$) + VADv2 ($\sigma=0.1$) & 1:1 & 0.423 & 0.699 & 0.496 & 0.285 & 0.741 & 0.181 & 0.149 \\
    M9 & VAD ($\sigma=0.2$) + Random & 5:1 & 0.474 & 0.609 & 0.533 & 0.322 & 0.713 & 0.175 & 0.151 \\
    M6 & Random & -- & 0.453 & 0.67 & 0.7 & 0.73 & 0.742 & 0.173 & 0.141 \\
    M19 & VAD ($\sigma=0.1$) + VAD ($\sigma=0.2$) + VADv2 ($\sigma=0.1$) & 1:1:1 & 0.423 & 0.65 & 0.467 & 0.203 & 0.742 & 0.168 & 0.149 \\
    M20 & IL & -- & 0.73 & 0.341 & 0.45 & 0.276 & 0.659 & 0.116 & 0.092 \\
    M10 & VADv2 ($\sigma=0.2$) & -- & 0.526 & 0.565 & 0.448 & 0.202 & 0.831 & 0.095 & 0.08 \\
    \bottomrule
  \end{tabular}
  }
  \label{tab:appendix_policy_mixtures}
\end{table*}

\section{Additional Qualitative Results}
\label{sec:appendix_qualitative}

\begin{figure*}[htbp]
  \centering
  \begin{subfigure}{0.48\linewidth}
    \includegraphics[width=0.98\linewidth]{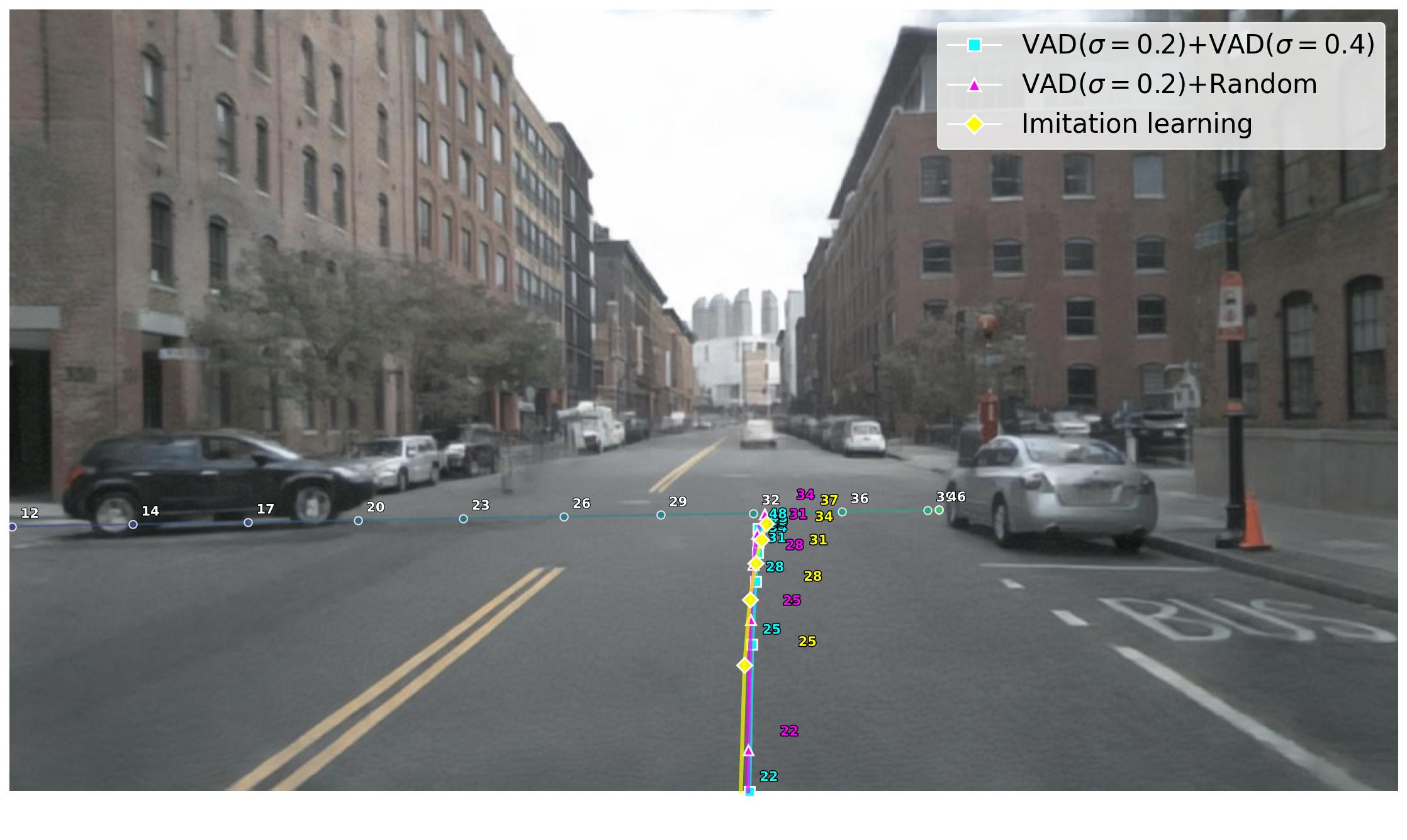}
    \caption{}
    \label{fig:app_qualitative_results_a}
  \end{subfigure}
  \hfill
  \begin{subfigure}{0.48\linewidth}
    \includegraphics[width=0.98\linewidth]{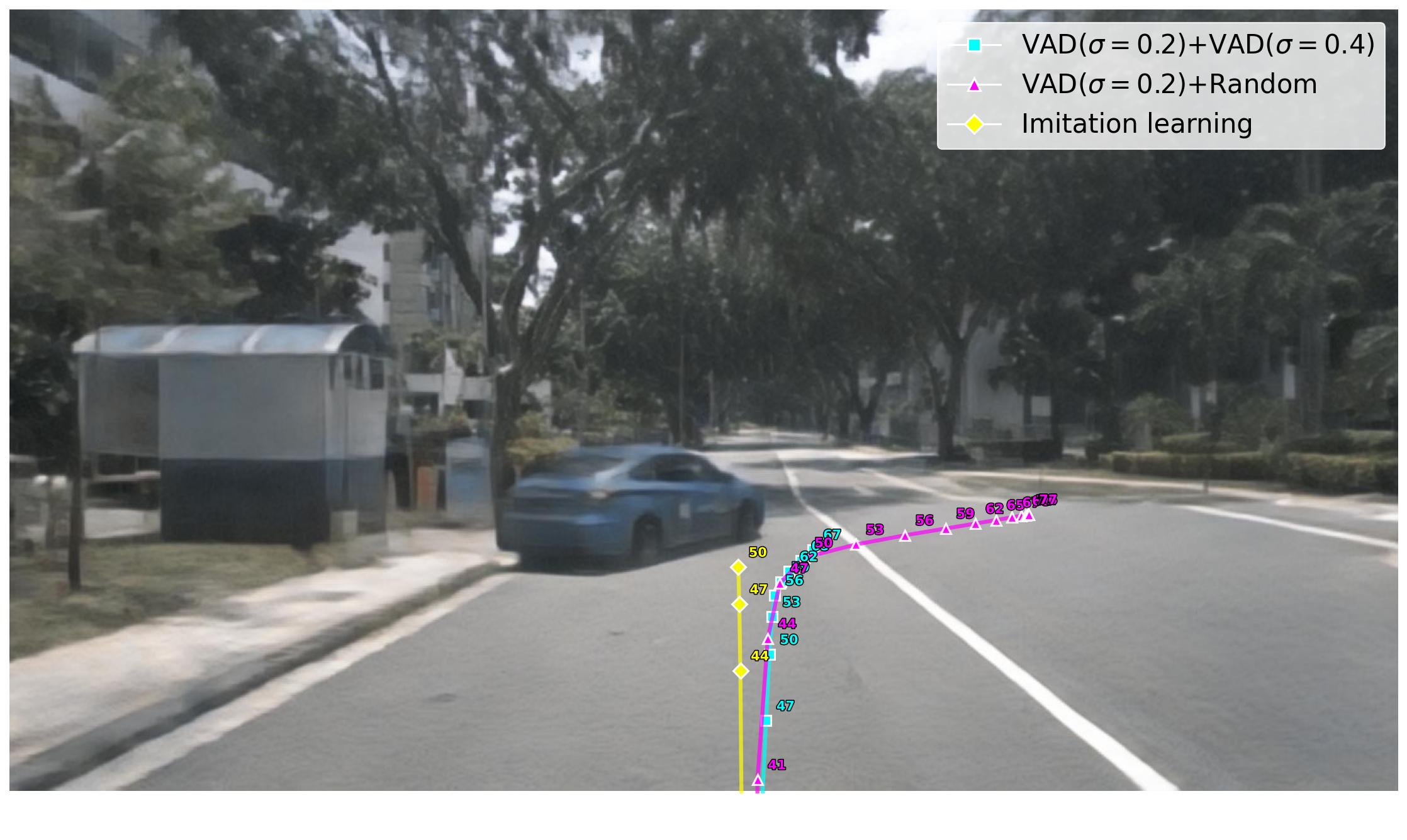}
    \caption{}
    \label{fig:app_qualitative_results_b}
  \end{subfigure} \\
    \begin{subfigure}{0.48\linewidth}
    \includegraphics[width=0.98\linewidth]{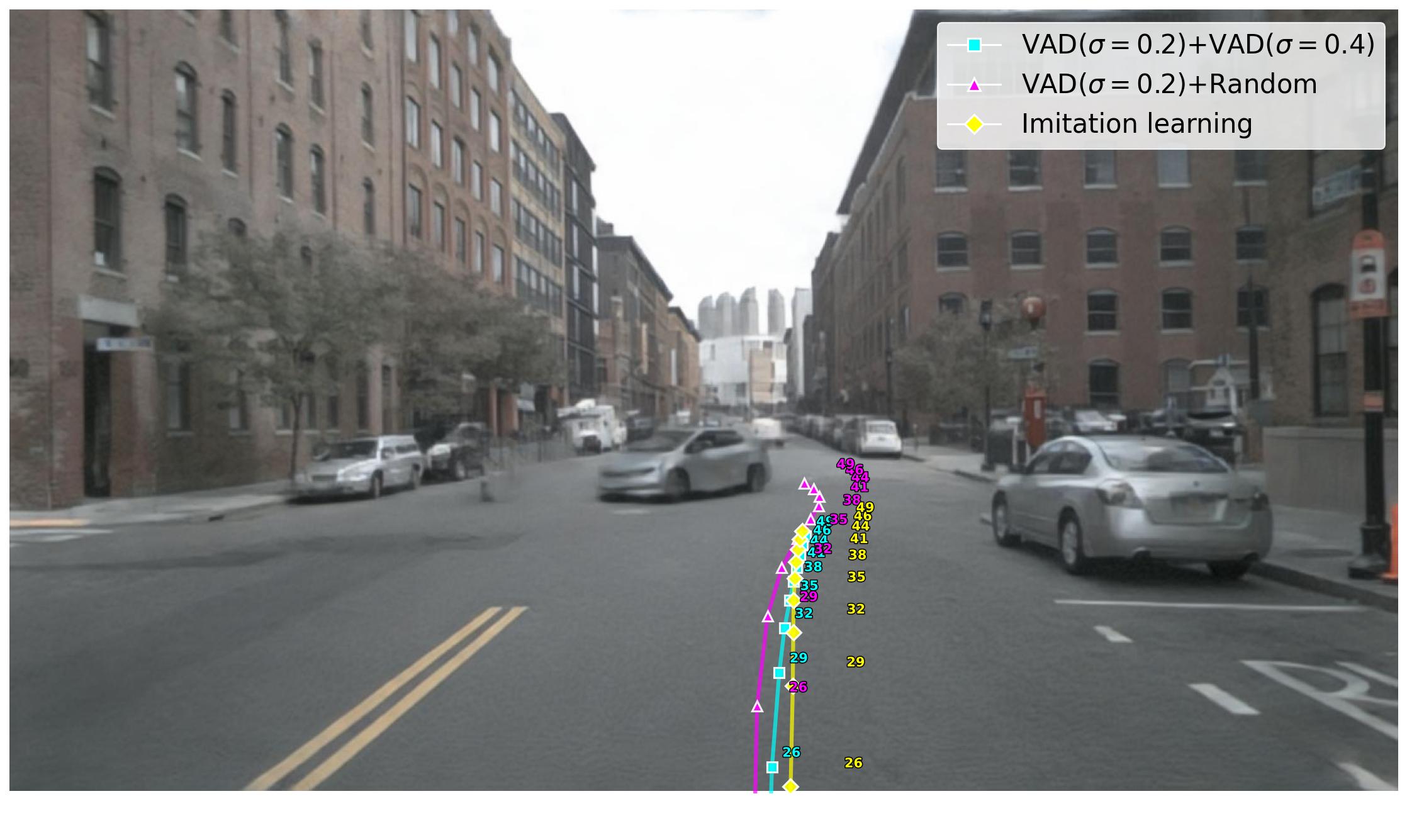}
    \caption{}
    \label{fig:app_qualitative_results_c}
  \end{subfigure}
  \hfill
  \begin{subfigure}{0.48\linewidth}
    \includegraphics[width=0.98\linewidth]{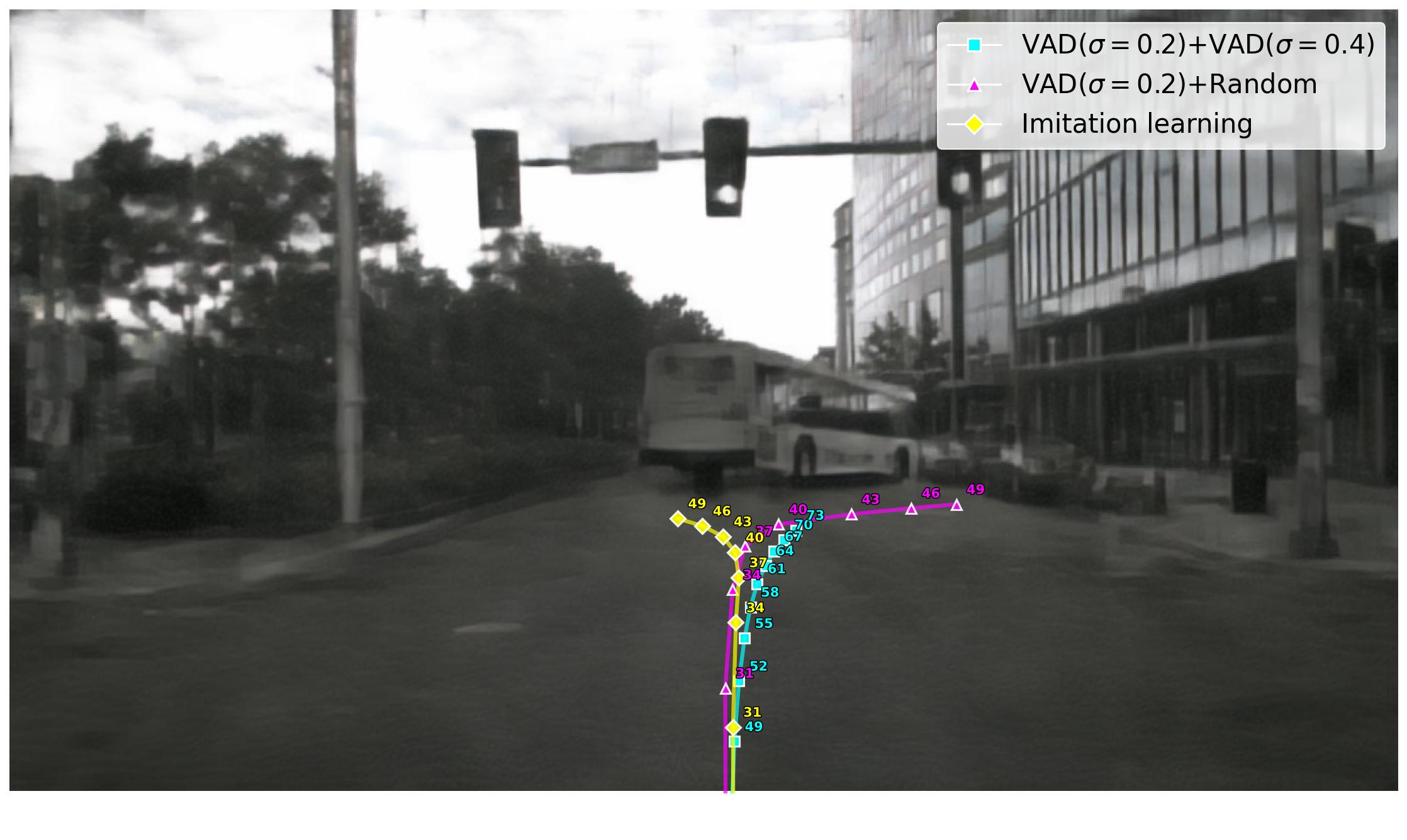}
    \caption{}
    \label{fig:app_qualitative_results_d}
  \end{subfigure} \\
  \begin{subfigure}{0.48\linewidth}
    \includegraphics[width=0.98\linewidth]{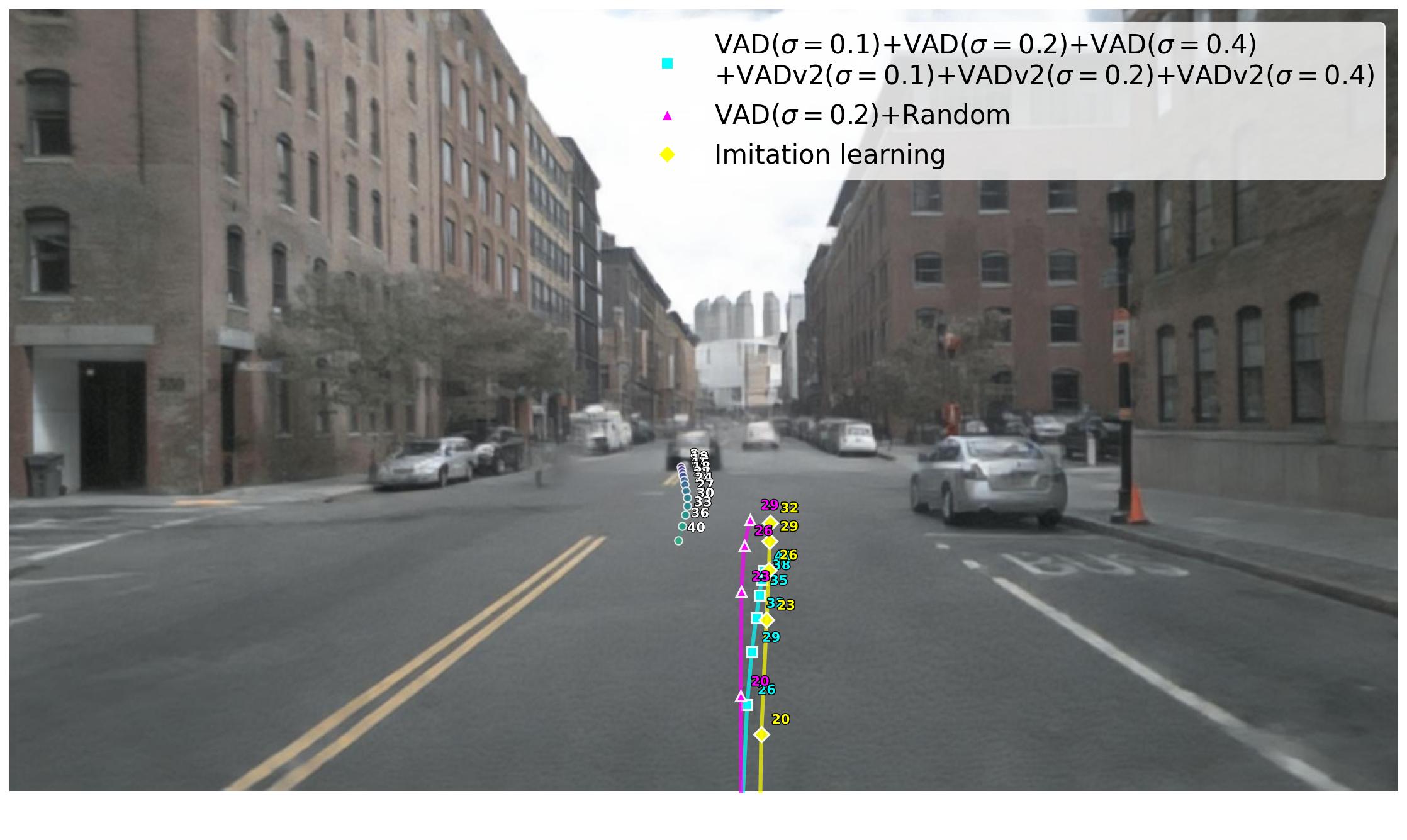}
    \caption{}
    \label{fig:app_qualitative_results_e}
  \end{subfigure}
  \hfill
  \begin{subfigure}{0.48\linewidth}
    \includegraphics[width=0.98\linewidth]{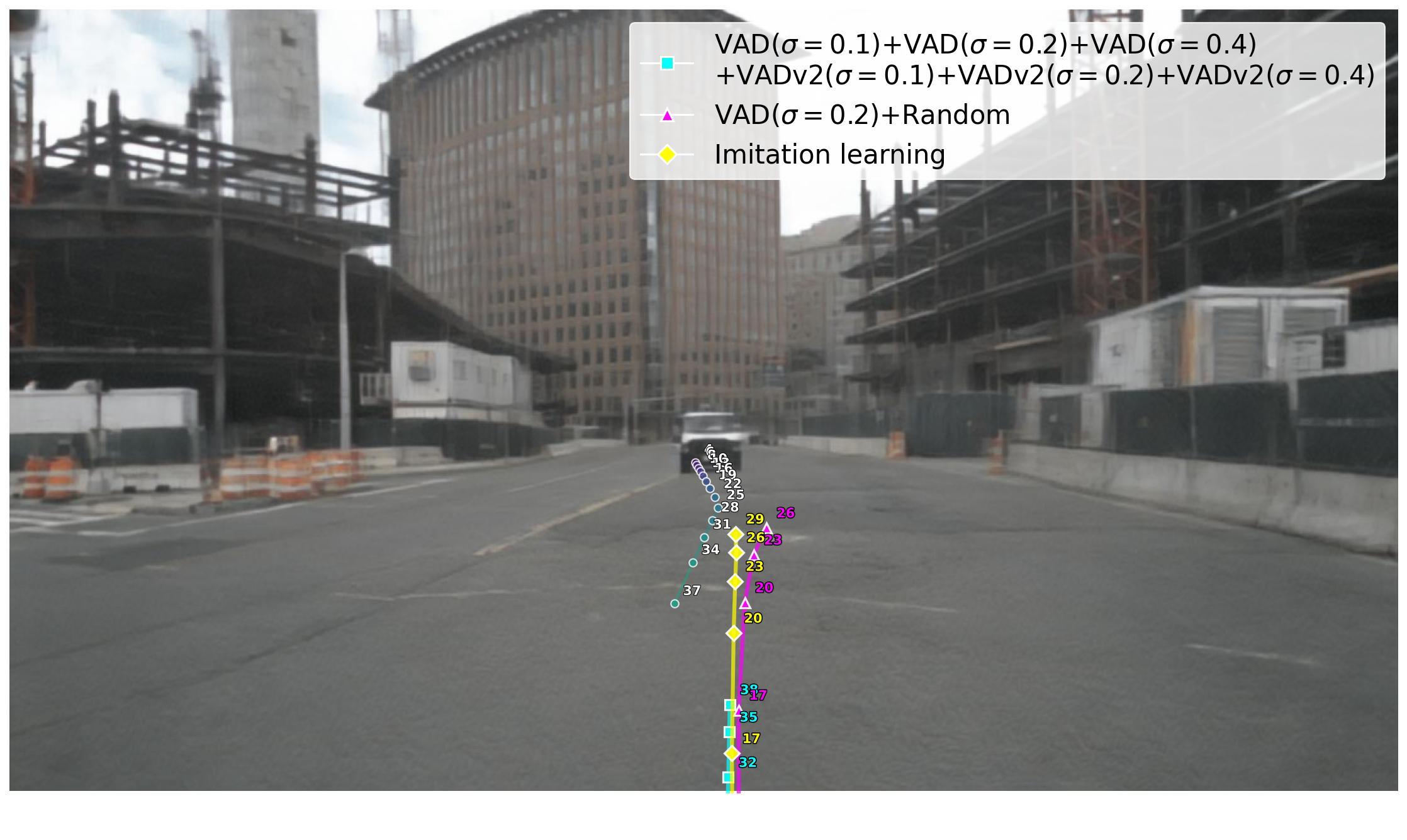}
    \caption{}
    \label{fig:app_qualitative_results_f}
  \end{subfigure}
  \caption{\textbf{Additional qualitative results in safety-critical NeuroNCAP scenarios.} We compare trajectories from the IL baseline, the VADv2† model (trained with Random data), and our VADv2* model. \textbf{(a)} An adversarial vehicle approaches from the left. \textbf{(b, c, d)} A stationary vehicle or bus obstructs the lane. \textbf{(e, f)} An adversarial vehicle approaches head-on. In (e, f), VADv2* is replaced with the VADv2‡ model (M14, a 6-policy mix).}
  \label{fig:app_qualitative_results}
\end{figure*}

To further illustrate the behavioral differences between the learned policies, Figure~\ref{fig:app_qualitative_results} provides additional qualitative examples from the safety-critical NeuroNCAP benchmark. We detail the behavior of the IL baseline, our offline RL model (VADv2*, trained on VAD($\sigma=0.2$) + VAD($\sigma=0.4$)), and the RL ablation (VADv2†, trained on VAD($\sigma=0.2$) + Random). As discussed in Sec.~\ref{sec:behavior_policy_and_learned_strategy}, VADv2* is a safety-focused policy, while VADv2† is an efficiency-focused policy. The scenarios below highlight these distinct driving `personalities'.

\begin{itemize}
    \item[\textbf{(a)}] \textbf{Adversarial vehicle from left:} The IL baseline and VADv2† fail to react to the approaching vehicle and collide. In contrast, VADv2* successfully identifies the hazard and avoids the collision by stopping short of the adversarial vehicle's trajectory.

    \item[\textbf{(b)}] \textbf{Stationary vehicle (center):} The IL baseline proceeds straight and collides with the obstacle. VADv2† successfully avoids a collision but does so with a high-risk, sharp swerve into the oncoming lane. VADv2* performs a safer, more controlled maneuver by slowing and navigating around the vehicle.

    \item[\textbf{(c)}] \textbf{Stationary vehicle (center):} In this scenario, all three models successfully avoid a collision. The IL baseline and VADv2* both slow down appropriately, while VADv2† again opts for a more aggressive swerving maneuver.

    \item[\textbf{(d)}] \textbf{Bus blocking lane:} A large bus completely blocks the road, requiring a full stop. The IL baseline and VADv2† both attempt an evasive maneuver but fail to stop, resulting in a collision. VADv2* correctly identifies the situation and comes to a safe stop before the obstacle.

    \item[\textbf{(e, f)}] \textbf{Frontal head-on collision:} These are among the most challenging scenarios, with the lowest success rates in the benchmark. Here, we show results for the VADv2‡ model (M14, trained on a 6-policy mix) instead of VADv2*. While all three models (IL, VADv2†, VADv2‡) ultimately fail to avoid a collision, VADv2‡ demonstrates a superior learned response by performing a minimal safety reaction and stopping just before impact.
\end{itemize}

\section{Training Stability Techniques}
\label{sec:appendix_training_stability}

Offline RL is prone to instability, particularly early in training when the critic is inaccurate. We use three standard stabilization techniques: (1) ramping up the actor-loss weight, (2) EMA target networks for TD target computation, and (3) a discrete action space.

\noindent
\textbf{Actor Loss Weight Scheduling.}
Applying the full actor loss from the start can destabilize training because the critic is still immature. We therefore use an exponential ramp-up schedule. The actor-loss weight is

$$w_{\text{actor}} = w_{\text{base}} \cdot \min\left(w_{\max}, w_{\max} \cdot w_{\text{init}} \cdot \rho^{t}\right),$$
where $t$ denotes the training iteration, $w_{\text{base}}$ is a global scale, $w_{\max}$ is the cap, $w_{\text{init}}$ is the initial value, and $\rho > 1$ is the growth rate. The scaled actor objective is
$$\mathcal{L^\prime}_{\text{actor}} = w_{\text{actor}} \cdot \mathcal{L}_{\text{actor}}.$$
In our experiments, $w_{\text{base}}=10$, $w_{\max}=1$, $w_{\text{init}}=10^{-4}$, and $\rho=1.0004$.

\noindent
\textbf{EMA Target Network.}
Following standard deep RL practice, we maintain slowly updated target copies of the online networks. Denoting online parameters by $\theta$ and target parameters by $\theta'$, we update the latter by EMA:
$$\theta'^{(t+1)} = (1-\tau) \theta'^{(t)} + \tau \theta^{(t)},$$
where $\tau \in [0,1)$ is the EMA coefficient. In our experiments, we use $\tau = 1 \times 10^{-4}$.
The TD target in the critic loss is computed from the target actor $\pi_{\theta'}$ and target critic $Q_{\psi'}$:
$$y = r + \gamma (1-d) Q_{\psi'}(s', a'), \quad a' \sim \pi_{\theta'}(\cdot \mid s'),$$
where $d$ is the terminal indicator.

\noindent
\textbf{Discrete Action Space.}
We also adopt a discrete action space. This makes the policy objective a categorical log-likelihood, so terms such as $\log \pi_\theta(a \mid s)$ are computed directly by a softmax classifier. In contrast, continuous-action policies typically require explicit density parameterization and more delicate log-likelihood computation. This simpler objective improves numerical stability in both the actor update and the pseudo-expert BC term.

\section{Data Collection Details}
\label{sec:appendix_data_collection}

The offline dataset is generated by running behavior policies in the NeuroNCAP simulator on scenes from the nuScenes training split. Each scene is approximately 20 seconds long. To increase coverage, we start rollouts at 3-second intervals from the beginning of each scene, yielding approximately six rollouts per scene. We refer to one complete pass over all training scenes as a \textit{sweep}. Each rollout terminates either when the scene ends or when a terminal event occurs (collision, off-road, or off-route). In the latter case, the terminal flag is set and used in the subsequent TD target computation.

All datasets used in our experiments consist of 12 sweeps. For mixed-policy datasets, the sweeps are allocated in proportion to the specified mixing ratio. For example, M2 in Table~\ref{tab:appendix_policy_mixtures} uses VAD ($\sigma=0.2$) + VAD ($\sigma=0.4$) with a 1:1 ratio, corresponding to 6 sweeps collected with VAD ($\sigma=0.2$) and 6 sweeps collected with VAD ($\sigma=0.4$).

The average rollout length varies across behavior policies because the frequency of early termination differs. For example, VAD ($\sigma=0.2$) yields an average of approximately 11.3 frames per rollout, whereas the Random policy yields approximately 7.6 frames. The shorter average rollout length of the Random policy reflects its higher likelihood of triggering terminal events due to uniformly random action selection.

\end{document}